\documentclass{article}

 \usepackage[main, final]{neurips_2026}

\newcommand{\ours}[1]{FocusFlow}
\usepackage[utf8]{inputenc} 
\usepackage[T1]{fontenc}    
\usepackage{hyperref}       
\usepackage{url}            
\usepackage{booktabs}       
\usepackage{amsfonts}       
\usepackage{nicefrac}       
\usepackage{microtype}      
\usepackage{xcolor}         

\usepackage{booktabs}        
\usepackage{threeparttable}  

\usepackage{graphicx}
\usepackage{amsmath}
\usepackage[linesnumbered,ruled,vlined]{algorithm2e}

\SetKwInput{KwIn}{Input}                
\SetKwInput{KwOut}{Output}              
\SetAlCapHSkip{0pt}                     

\title{Training-Free 3D Editing via Feature-Divergence Localization and Trajectory Correction}

\author{%
  Tri Ton \qquad Siwoo Lim \qquad Joshua Tian Jin Tee \qquad Chang D.~Yoo\thanks{Corresponding Author} \\
  Korea Advanced Institute of Science and Technology (KAIST) \\
  \texttt{\{tritth, cd\_yoo\}@kaist.ac.kr}
}

\begin{document}

\maketitle

\begin{abstract}
Modifying 3D assets requires enacting targeted structural transformations while strictly preserving the geometric identity of unedited regions. While recent probability flow models enable training-free editing, bounding these modifications within a high-dimensional latent space presents severe mathematical challenges. Current inversion-free trajectories rely on global velocity updates, which often cause spatial bleed into preserved structures and induce trajectory under-commitment, causing edits to stall mid-path. To resolve these dual failures, we introduce \ours{}, a segmentation-free framework that structures the integration trajectory into two disjoint phases. During early coarse structure formation, \ours{} applies Feature-Divergence Localization (FDL) to spatially gate the velocity field using intrinsic cross-attention. Once coarse structure has formed, \ours{} applies Reference Generation Guidance (RGG), a trajectory-correction term that drives convergence to the target shape by pulling the trajectory toward an explicit clean-latent target. 
Evaluated on the joint Google Scanned Objects and PartObjaverse-Tiny benchmarks, \ours{} obtains the best preservation scores and the largest edit magnitude among compared methods, suggesting that it can reduce the traditional preservation-modification trade-off.
By providing stable, localized geometric control without manual spatial annotations, this phase-scheduled approach lowers technical barriers to the modification of synthetic 3D media.
\end{abstract}    

\section{Introduction}
\label{sec:intro}

Modifying existing 3D assets~\cite{bar2025editp23, ye2025nano3d, zhou2025anchorflow} requires enacting targeted structural transformations while strictly preserving the geometric identity of unedited regions. Recent 3D foundation models~\cite{hunyuan3d2025hunyuan3d, xiang2025structured} parameterized as rectified flow transformers provide strong geometric priors for shape generation. However, extending these priors to training-free editing via inversion-free~\cite{kulikov2025flowedit, zhou2025anchorflow, kim2025flowalign, ye2025nano3d} trajectories introduces severe failure modes. Specifically, current unconstrained editing trajectories either stall before realizing the target geometry or incorrectly propagate the transformation into regions that must remain identical to the source shape.

\begin{figure}[t!]
  \centering
  \includegraphics[width=.7\linewidth]{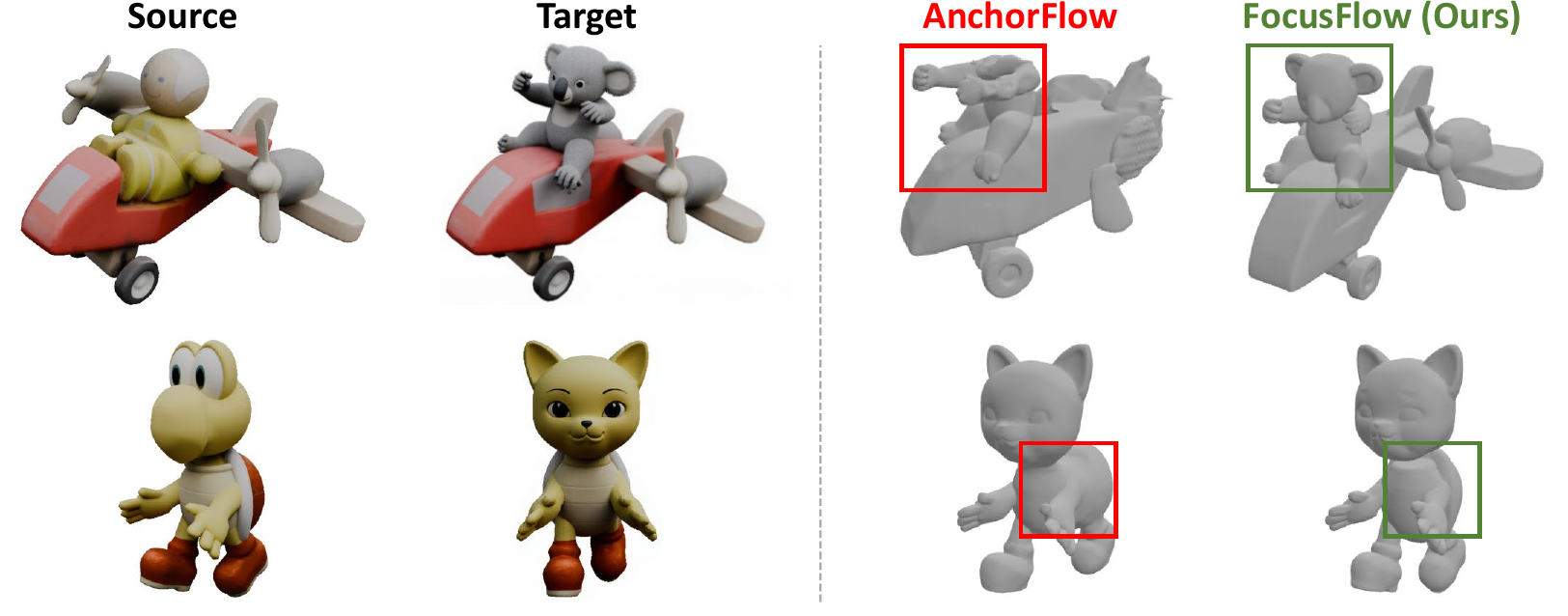}
  \caption{\textbf{\ours{} addresses global editing failures.} The baseline~\cite{zhou2025anchorflow} exhibits spatial bleed and trajectory under-commitment (red boxes). \ours{} confines the velocity field during localization and pulls toward a clean-latent target during guidance, realizing the target with less background distortion and no segmentation masks (green boxes).}
  \label{fig:teaser}
  \vskip -0.15in
\end{figure}

These failures originate from the underlying dynamics of probability flow ordinary differential equations (ODEs). Rectified flows establish deterministic transport maps designed to follow straight paths between empirical distributions. Because global editing updates, such as AnchorFlow~\cite{zhou2025anchorflow}, apply a uniform semantic velocity across the entire latent representation, structures that should remain unchanged absorb part of the spatial transformation. As shown in Fig.~\ref{fig:teaser} (top), this results in severe spatial bleed where the plane's structure is distorted even though the edit targets only the passenger. Furthermore, flow trajectories driven strictly by velocity differences encode a semantic direction of change but lack an explicit terminal coordinate. Consequently, edits of medium magnitude suffer from trajectory under-commitment in Fig.~\ref{fig:teaser} (bottom), stalling mid-path and failing to physically realize the target geometry. Naive attempts to resolve this stalling via instantaneous target activation abruptly shift the velocity field, destabilizing the integration path and inducing topological fracturing (the mesh tearing into disconnected fragments) in the decoded mesh.

To resolve these dual failures, we introduce \ours{} (Fig.~\ref{fig:teaser}, right), a phase-scheduled editing framework that strictly decouples spatial gating from terminal convergence. Prior studies in 2D image manipulation demonstrate that edits can be localized without explicit manual annotations, using cross-attention maps~\cite{hertz2022prompt} or contrasts between noise predictions~\cite{couairon2022diffedit}. 
Building on this observation, \ours{} integrates Feature-Divergence Localization (FDL) to enforce intrinsic spatial localization and prevent spatial bleed, alongside Reference Generation Guidance (RGG), a trajectory-correction term that drives explicit endpoint commitment. By scheduling these mechanisms into sequential, disjoint phases of the integration trajectory, \ours{} avoids the weaker edits we observe when both act at once.
Our core technical contributions are threefold. 
First, we propose a phase-scheduled integration that assigns spatial gating and terminal convergence to disjoint phases of the trajectory, avoiding the weaker edits we observe when both act at once.
Second, for intrinsic spatial localization, we propose Feature-Divergence Localization (FDL), a segmentation-free module deriving a soft gate from patch divergence and cross-attention scores to suppress spatial bleed during early integration. Third, to encourage explicit endpoint commitment, we introduce Reference Generation Guidance (RGG), which smoothly pulls the trajectory toward a clean-latent reference target via time-dependent weighting, reducing the risk of topological fracturing. On the Google Scanned Objects (GSO)~\cite{downs2022google} and PartObjaverse-Tiny~\cite{yang2024sampart3d} benchmarks, \ours{} obtains the best preservation scores and the largest edit magnitude among compared methods.

\section{Related Work}
\label{sec:related}

\paragraph{3D generation models.}
Recent 3D generative models improve the quality and accessibility of digital asset creation. Early optimization-based approaches demonstrated that pretrained 2D diffusion priors~\cite{rombach2022high} can be distilled into 3D representations through rendered-view supervision~\cite{poole2022dreamfusion, lin2023magic3d}. However, these methods require costly per-instance optimization and are designed for generating new assets rather than modifying existing ones. More recent native 3D generative models~\cite{hunyuan3d2025hunyuan3d, xiang2025structured, zhang2024clay, jun2023shap, wu2024direct3d, luo2021diffusion, nichol2022point} learn generative priors directly over large-scale 3D data. Specific architectures predict 3D representations from a single image~\cite{hong2023lrm}, introduce structured 3D latents decodable into multiple formats~\cite{xiang2025structured}, or provide high-fidelity image-to-3D pipelines based on learned shape latents~\cite{hunyuan3d2025hunyuan3d}. While these models provide strong geometric priors, they function as generators rather than editors. They synthesize new assets but lack an explicit mechanism to preserve a source asset while modifying only a localized region. 

\paragraph{Training-free 3D editing.}

Existing 3D editing methods balance localized modifications with source geometry preservation. Classical approaches deform or reconstruct meshes under semantic guidance via text-driven differentiable rendering~\cite{gao2023textdeformer}, conditional reconstruction~\cite{gao20253d}, region-level fusion~\cite{jincheng2025craftmesh}, or part-aware pipelines~\cite{liu2024part123, yan2025x}, but rely on predefined editable regions, segmentation cues, or auxiliary supervision. Recent work instead edits assets in native latent spaces to leverage geometric priors and avoid multi-view inconsistencies~\cite{li2025voxhammer, ye2025nano3d, parelli20253d}, with flow-based editors using velocity differences~\cite{kulikov2025flowedit} or latent anchors~\cite{zhou2025anchorflow}, and feed-forward variants enabling fast edits~\cite{hu2026easy3e}. Still, these face a strict preservation-modification trade-off, and training-free variants either entangle geometry with texture~\cite{li2025voxhammer, ye2025nano3d} or require 3D masks~\cite{li2025voxhammer}, drawn by hand or predicted by 3D segmentation models~\cite{yang2024sampart3d, ton2024dual}. 
In 2D, Prompt-to-Prompt~\cite{hertz2022prompt} shows that cross-attention ties prompt words to image regions, and DiffEdit~\cite{couairon2022diffedit} derives an edit mask by contrasting noise predictions under the two prompts. The 3D latent is an unordered token set with no spatial grid, so FDL instead maps DINO patch differences to latent tokens through the generator’s cross-attention.
\ours{} operates strictly on the geometric latent and separates spatial localization from target commitment, enabling precise mask-free editing.

\section{Preliminaries}
\label{sec:prelim}

\subsection{Inversion-Free 3D Editing}
\label{sec:prelim:flow}

We build on a rectified flow transformer~\cite{hunyuan3d2025hunyuan3d, lipman2022flow,liu2022flow, esser2024scaling}
that maps a Gaussian prior to a 3D shape latent distribution via a learned velocity field. 
Let $z_t \in \mathbb{R}^{N \times d}$ denote the latent state at flow time $t \in [0,1]$, 
where $N$ represents the token count and $d$ represents the channel dimension. The generator parameterizes a conditional velocity field $v_\theta(z_t, t, c)$, driven by an image condition $c$. The probability flow ordinary differential equation (ODE) trajectory is defined as $\mathrm{d}z_t \;=\; v_\theta(z_t, t, c)\mathrm{d}t$.
This equation governs the deterministic transport of an initial Gaussian sample $z_0 \sim \mathcal{N}(0, I)$ to the clean 3D shape latent $z_1$.
Integration of this vector field across the unit interval yields a sample from the learned shape distribution.

To edit a 3D shape, the system receives a source condition $c^{\mathrm{src}}$ and a target condition $c^{\mathrm{tar}}$. The objective is to compute an edited trajectory $\{z_t^{\mathrm{edit}}\}_{t \in [0,1]}$ that originates at the source clean latent $z_1^{\mathrm{src}}$ and evolves into an edited latent $z_1^{\mathrm{tar}}$. We adopt the inversion-free formulation introduced by FlowEdit~\cite{kulikov2025flowedit}, which defines the edited trajectory dynamically as
\begin{equation}
\label{eq:edit_traj}
z_t^{\mathrm{edit}} \;=\; z_1^{\mathrm{src}} \;+\; z_t^{\mathrm{tar}} \;-\; z_t^{\mathrm{src}}.
\end{equation}
Here $z_t^{\mathrm{src}} = t\,z_1^{\mathrm{src}} + (1-t)\,\epsilon$ with $\epsilon \sim \mathcal{N}(0, I)$ is the source latent noised to level $t$, and $z_t^{\mathrm{tar}} = z_t^{\mathrm{edit}} + z_t^{\mathrm{src}} - z_1^{\mathrm{src}}$
follows from Equation~\ref{eq:edit_traj}.
This formulation constructs a path that begins at the source shape and ends at the edited shape, avoiding the destructive process of transporting the latent back through the Gaussian prior space. By defining $v_t^{\mathrm{src}} = v_\theta(z_t^{\mathrm{src}}, t, c^{\mathrm{src}})$ and $v_t^{\mathrm{tar}} = v_\theta(z_t^{\mathrm{tar}}, t, c^{\mathrm{tar}})$, we obtain the inversion-free edit velocity by differentiating Equation~\ref{eq:edit_traj} with respect to $t$: $v_t^{\mathrm{edit}} \;=\; v_t^{\mathrm{tar}} \;-\; v_t^{\mathrm{src}}.$ 

\subsection{Latent Anchor-Aligned Flows}
\label{sec:prelim:anchor}

While inversion-free editing prevents Gaussian prior destruction, unconstrained velocity differences frequently push the latent representation off the target data manifold. AnchorFlow~\cite{zhou2025anchorflow} constrains this geometric drift by introducing a latent-anchor alignment term. 
The latent anchor $f_t$ is defined as
$f_t \;=\; z_t \;+\; (1 - t)v_t.$
The anchor $f_t$ computes the linear forecast of the clean latent by extending the current velocity tensor directly to the endpoint of the trajectory. To enforce proximity between the source and target trajectories in this forecasted clean-latent space, AnchorFlow derives an alignment loss $\mathcal{L}_{\mathrm{align}}$. The gradient descent step on this alignment loss evaluates to
\begin{equation}
\label{eq:lalign}
\nabla \mathcal{L}_{\mathrm{align}} \;=\; (2 - t)\big(f_t^{\mathrm{tar}} \;-\; f_t^{\mathrm{src}}\big).
\end{equation}
This gradient step actively penalizes geometric divergence between the two forecasts, pulling the edit trajectory back toward the structurally stable source manifold. The complete, unconstrained AnchorFlow update is therefore applied at each integration step as
{
\begin{equation}
\label{eq:anchorflow}
\mathrm{d}z_t^{\mathrm{edit}} \;=\; \big(v_t^{\mathrm{edit}} \;-\; \nabla \mathcal{L}_{\mathrm{align}}\big)\mathrm{d}t.
\end{equation}
}

While Equation~\ref{eq:anchorflow} preserves manifold adherence, applying this update globally across all $N$ tokens causes severe \textit{spatial bleed}, altering background structures that must remain preserved. Furthermore, the velocity vector $v_t^{\mathrm{edit}}$ directs semantic change but lacks an explicit terminal coordinate. This absence causes \textit{trajectory under-commitment}, where medium-magnitude edits stall mid-path and fail to realize the target shape.

\section{Method}
\label{sec:method}

\subsection{Phase-Scheduled Framework}
\label{sec:method:arch}

We distinguish continuous flow time $t \in [0, 1]$ from the edit-step index $i = 0, 1, \dots$, which counts solver steps from the start of the edit. While unconstrained velocity integration theoretically maps the source distribution to the target distribution~\cite{kulikov2025flowedit}, in practice it often stalls mid-path~\cite{zhou2025anchorflow}: the velocity difference encodes a direction of change but no explicit terminal coordinate. Furthermore, because flow models form global layouts under high noise and refine details under low noise, spatial gating should act during early structural formation, whereas pulling toward a terminal target only helps once coarse geometry exists. To resolve spatial bleed and trajectory under-commitment, \ours{} structures the integration trajectory into two disjoint phases separated by a discrete boundary $s^{\mathrm{mask}}$ ($5$ steps in our experiments), decoupling early spatial localization from subsequent terminal convergence. Figure~\ref{fig:arch} illustrates the complete phase-scheduled framework:

\begin{figure}[t!]
  \centering
  \includegraphics[width=\linewidth]{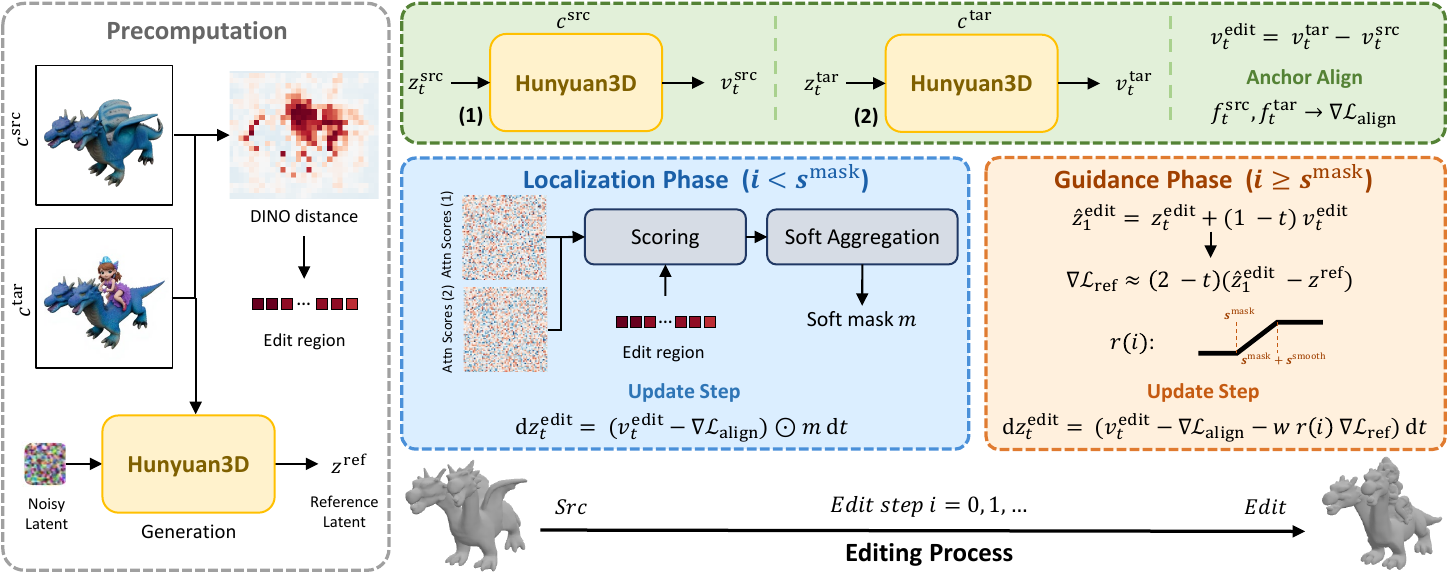}
  \caption{\textbf{\ours{} framework.} \emph{Left:} Precomputation of the DINO distance, the edit region and the reference latent $z^{\mathrm{ref}}$. \emph{Top:} Per-step velocities and the alignment gradient. \emph{Center (localization phase):} FDL gates the velocity field via soft mask $m$ to confine the edit. \emph{Right (guidance phase):} RGG applies a ramped pull toward $z^{\mathrm{ref}}$, driving structural commitment.}
  \label{fig:arch}
  \vskip -0.15in
\end{figure}

\begin{itemize}
    \item \textbf{Localization Phase ($i < s^{\mathrm{mask}}$):} During the early integration steps, Feature-Divergence Localization (FDL) constructs a dynamic per-token soft mask $m$ to spatially gate the velocity field, suppressing spatial bleed into preserved regions.
    \item \textbf{Guidance Phase ($i \ge s^{\mathrm{mask}}$):} Once coarse geometry forms, the spatial gate retires. Reference Generation Guidance (RGG) activates via a weight ramped over the step index, pulling the trajectory toward an explicit clean-latent target to drive convergence to the target shape.
\end{itemize}
We detail the spatial gating and target convergence mechanisms below.

\subsection{Feature-Divergence Localization}
\label{sec:method:fdl}

To prevent spatial bleed during coarse-structure formation, Feature-Divergence Localization (FDL) generates a continuous soft mask $m$ that spatially gates the velocity field. To maintain structural stability, we derive this mask from the divergence of semantic image features rather than the divergence of the high-dimensional velocity field. We construct this intrinsic spatial localization mechanism through three sequential operations: semantic patch selection, multi-block cross-attention scoring, and soft aggregation.

\paragraph{Semantic Patch Selection.}
Let $c^{\mathrm{src}}$ and $c^{\mathrm{tar}}$ denote the source and target image conditions. 
Let $\mathcal{E}_{\mathrm{DINO}}$ denote a frozen DINO encoder~\cite{oquab2023dinov2}. We pass each image condition through $\mathcal{E}_{\mathrm{DINO}}$ to extract per-patch features $\phi^{\mathrm{src}} = \mathcal{E}_{\mathrm{DINO}}(c^{\mathrm{src}})$ and $\phi^{\mathrm{tar}} = \mathcal{E}_{\mathrm{DINO}}(c^{\mathrm{tar}})$, both in $\mathbb{R}^{P \times d_{\!\phi}}$, where $P$ is the total number of patches and $d_{\!\phi}$ is the feature dimension. We compute the patch-wise cosine distance to measure the semantic deviation:
\begin{equation}
\label{eq:dino}
D_p \;=\; 1 \;-\; \frac{\phi^{\mathrm{src}}_p \cdot \phi^{\mathrm{tar}}_p}{\| \phi^{\mathrm{src}}_p \| \, \| \phi^{\mathrm{tar}}_p \|}, \qquad p \in \{1, \dots, P\}.
\end{equation}
Because DINO features encode semantic correspondences rather than low-level pixel intensities, patches exhibiting high cosine distance likely reflect semantic changes. Instead of selecting a fixed number of patches, we dynamically define the edit region $K_{\mathrm{changed}} \subseteq \{1,\dots,P\}$ using an adaptive statistical threshold. 
We retain all patches where $D_p > \mu(D) + \sigma(D)$, where $\mu(D)$ and $\sigma(D)$ are the mean and standard deviation of $D$ over the $P$ patches. This yields a sparse edit region whose size adapts to the edit.

\paragraph{Multi-Block Cross-Attention Scoring.}
To translate this image-space divergence into a latent-space relevance score, we intercept the cross-attention queries $q_b \in \mathbb{R}^{N \times d_k}$ from latent tokens and keys $k_b \in \mathbb{R}^{P \times d_k}$ from image patches. We perform this interception across a selected subset of transformer blocks $\mathcal{B}$. For each block $b \in \mathcal{B}$ and latent token $n$, we compute the unnormalized relevance score by averaging the dot product strictly over the changed patches:
\begin{equation}
\label{eq:attn_score}
a_{b,n} \;=\; \frac{1}{|K_{\mathrm{changed}}|} \sum_{p \in K_{\mathrm{changed}}} q_{b,n} \cdot k_{b,p}.
\end{equation}

We compute these scores independently for the source and target forward passes and min-max normalize each over the $N$ latent tokens, so that neither pass dominates the fusion through its raw scale. This yields $\bar{a}_{b,n}^{\mathrm{src}}, \bar{a}_{b,n}^{\mathrm{tar}} \in [0,1]$. To ensure that tokens relevant to either the original structure or the target geometry remain active, we fuse the two streams using an element-wise maximum:
$\mathrm{score}_{b,n} \;=\; \max\!\big(\bar{a}_{b,n}^{\mathrm{src}},\, \bar{a}_{b,n}^{\mathrm{tar}}\big).$

\begin{figure}[t!]
  \centering
  \includegraphics[width=.8\linewidth]{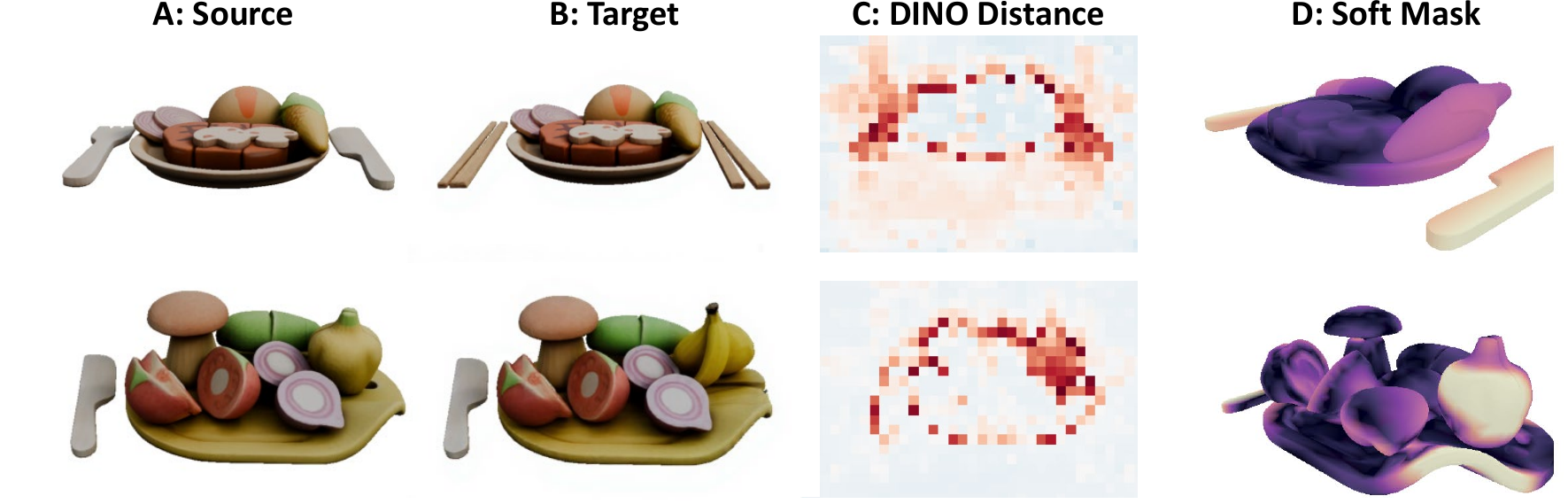}
  \caption{\textbf{FDL soft mask construction.} \emph{A:} Source condition. \emph{B:} Target condition. \emph{C:} DINO distance $D_p$ highlighting the semantic modification. \emph{D:} Final soft mask $m_n$. 
  }
  \label{fig:fdl_mask}
  \vskip -0.15in
\end{figure}

\paragraph{Soft Aggregation.}
The fused scores define a soft mask for each block $b$:
\begin{equation}
\label{eq:mask}
m_{b,n} \;=\; \alpha \;+\; (1 - \alpha)\,\bigl(\mathrm{score}_{b,n}\bigr)^{\gamma},
\end{equation}
where the floor $\alpha \in [0,1]$ is the minimum fraction of the update that every token receives, while the sharpness exponent $\gamma > 0$ controls the response boundary. 
We aggregate the block-wise masks via a mean operation to yield the final token mask $m_n = \frac{1}{|\mathcal{B}|} \sum_{b \in \mathcal{B}} m_{b,n}$, visualized in Fig.~\ref{fig:fdl_mask}.
We apply this intrinsic spatial localization during the localization phase ($i < s^{\mathrm{mask}}$):
{
\begin{equation}
\label{eq:gated}
\mathrm{d}z_t^{\mathrm{edit}} \;=\; \big(v_t^{\mathrm{edit}} \;-\; \nabla \mathcal{L}_{\mathrm{align}}\big) \odot m \, \mathrm{d}t.
\end{equation}
}

The operator $\odot$ denotes element-wise multiplication broadcast across the channels. 
We restrict gating to $i < s^{\mathrm{mask}}$ because flow models establish coarse geometry early and refine fine detail late, so the gate is needed only while coarse geometry forms. Moving the boundary later degrades both preservation and edit strength, whereas retiring the gate at $s^{\mathrm{mask}}$ lets the late steps update all tokens.

\subsection{Reference Generation Guidance} 
\label{sec:method:rgg}

While intrinsic spatial localization restricts the velocity domain, the differential signal $v_t^{\mathrm{edit}}$ strictly encodes a direction of semantic change, causing trajectories to stall before physically realizing the target shape. Reference Generation Guidance (RGG) resolves this trajectory under-commitment by establishing an explicit terminal coordinate in the continuous latent space. 
By precomputing a reference latent $z^{\mathrm{ref}}$ via the base flow generator~\cite{hunyuan3d2025hunyuan3d} on the target condition, we obtain a sample from the generator's prior that drives convergence to the target shape while bypassing the computational overhead and perspective biases of pixel-space differentiable rendering.

\paragraph{Clean Latent Residual.}

Since $z^{\mathrm{ref}}$ is a clean latent, we compare it with a forecast of the clean edited latent rather than with the current state $z_t^{\mathrm{edit}}$. This state is not a noisy sample at level $t$, so evaluating $v_\theta$ on it would query the network off its input distribution. We therefore form the forecast with the edit velocity $v_t^{\mathrm{edit}}$, which the network computes only at the noisy states $z_t^{\mathrm{src}}$ and $z_t^{\mathrm{tar}}$:
{
\begin{equation}
\label{eq:x0}
\hat{z}_1^{\mathrm{edit}} \;=\; z_t^{\mathrm{edit}} \;+\; (1-t)v_t^{\mathrm{edit}}.
\end{equation}
}

We enforce convergence by defining the clean latent reference loss as the squared Euclidean distance between this forecast and the reference latent:
\begin{equation}
\label{eq:lref}
\mathcal{L}_{\mathrm{ref}} \;=\; \frac{1}{2} \big\| \hat{z}_1^{\mathrm{edit}} \;-\; z^{\mathrm{ref}} \big\|^2.
\end{equation}
Differentiating $\mathcal{L}_{\mathrm{ref}}$ with respect to the current state $z_t^{\mathrm{edit}}$ requires applying the chain rule to Equation~\ref{eq:x0}, yielding an expression dependent on the target velocity Jacobian $J_\theta$:
\begin{equation}
\label{eq:lref-grad}
\nabla_{z_t^{\mathrm{edit}}} \mathcal{L}_{\mathrm{ref}} \;=\; \big(I \;+\; (1-t)J_\theta^\top\big) \big(\hat{z}_1^{\mathrm{edit}} \;-\; z^{\mathrm{ref}}\big),
\end{equation}
where $J_\theta = \partial v_\theta(z_t^{\mathrm{tar}}, t, c^{\mathrm{tar}}) / \partial z_t^{\mathrm{edit}}$.
The exact gradient needs a vector-Jacobian product at every guidance step. We substitute the exact matrix with the scalar identity $(2-t)I$, a standard simplification in the score-distillation literature~\cite{poole2022dreamfusion}. This substitution provides a directional pull toward the reference latent and yields a tractable, closed-form gradient:
\begin{equation}
\label{eq:lref-approx}
\nabla \mathcal{L}_{\mathrm{ref}} \;\approx\; (2-t)\big(\hat{z}_1^{\mathrm{edit}} \;-\; z^{\mathrm{ref}}\big).
\end{equation}
To prevent this residual from overriding the differential edit velocity and collapsing the geometry into a direct copy of $z^{\mathrm{ref}}$, we scale the gradient by a fixed balancing coefficient $w$. The applied guidance term becomes $w(2-t)(\hat{z}_1^{\mathrm{edit}} - z^{\mathrm{ref}})$.

\paragraph{Time-Dependent Weighting.}
Activating this reference guidance instantaneously at the phase boundary $s^{\mathrm{mask}}$ abruptly shifts the velocity field. This sudden transition destabilizes the integration, inducing topological fracturing in the decoded mesh. To ensure trajectory smoothing, we apply a continuous time-dependent weighting schedule $r(i)$ tied to the discrete step index, which scales linearly over a defined duration $s^{\mathrm{smooth}}$ ($15$ steps in our experiments):
$r(i) \;=\; \min\!\Big(1,\; \max\!\Big(0,\; \frac{i \;-\; s^{\mathrm{mask}}}{s^{\mathrm{smooth}}}\Big)\Big).$

Once the localization phase concludes and the spatial gate retires, this smoothed pull is introduced alongside the manifold alignment constraint. To minimize the distance loss and drive convergence toward the reference latent, we subtract the reference gradient:
{
\begin{equation}
\label{eq:guided}
\mathrm{d}z_t^{\mathrm{edit}} \;=\; \Big(v_t^{\mathrm{edit}} \;-\; \nabla \mathcal{L}_{\mathrm{align}} \;-\; w\ r(i)\nabla \mathcal{L}_{\mathrm{ref}}\Big)\mathrm{d}t.
\end{equation}
}

Applying this pull concurrently with the spatial gate during the early integration phase works against the edit: in our \textit{Concurrent} ablation (Table~\ref{tab:abl-component}), the background stays close to the source, but the edit is the weakest of all variants. Structuring the update into sequential, disjoint phases encourages stronger terminal convergence without violating the localized geometric structure.
The complete step-by-step pseudo-code for the phase-scheduled integration is provided in Appendix~\ref{sec:supp:algo}.

\section{Experiments}
\label{sec:exp}

\subsection{Experimental Setup}
\label{sec:exp:setup}

\begin{table}[t]
  \centering
  \caption{Quantitative comparison on GSO and PartObjaverse-Tiny. We mark the \textbf{best} and \underline{second-best} results. \textit{Note:} All reported numbers are scaled by a factor of $10^2$.}
  \label{tab:main}
  \small
  \scalebox{0.95}{
  \begin{tabular}{lcccccc}
    \toprule
    Method
      & $\mathrm{CD}_{\mathrm{preserve}}\!\downarrow$
      & $\mathrm{CD}_{\mathrm{edit}}\!\uparrow$
      & $\mathrm{NC}_{\mathrm{preserve}}\!\uparrow$
      & $\mathrm{F\text{-}score}\!\uparrow$
      & $\mathrm{CLIP}_{\mathrm{img}}\!\uparrow$
      & $\mathrm{CLIP}_{\mathrm{txt}}\!\uparrow$ \\
    \midrule
Direct Editing~\cite{hunyuan3d2025hunyuan3d}
      & 4.66 & 1.49 & 54.48 & \underline{2.98} & 63.30 & 23.64 \\
    Inversion~\cite{jiao2025uniedit}
      & 3.94 & 1.38 & 55.51 & 2.88 & \textbf{64.35} & \textbf{24.14} \\
    Inversion-free~\cite{kulikov2025flowedit}
      & \underline{3.15} & 1.50 & \underline{57.96} & 2.33 & 61.78 & 21.72 \\
    AnchorFlow~\cite{zhou2025anchorflow}
      & 3.81 & \underline{1.58} & 56.20 & 2.60 & 61.42 & 22.51 \\
    Nano3D~\cite{ye2025nano3d}
      & 3.22 & 1.50 & 57.46 & 2.50 & 62.06 & 22.08 \\
    \midrule
    \textbf{\ours{} (Ours)}
      & \textbf{3.11} & \textbf{1.61} & \textbf{58.20}
      & \textbf{3.05} & \underline{63.85} & \underline{23.91} \\
    \bottomrule
  \end{tabular}
  }
  \vskip -0.15in
\end{table}

\paragraph{Datasets.}
We evaluate on Edit3D-Bench~\cite{li2025voxhammer}, which contains  100 3D models curated from Google Scanned Objects (GSO)~\cite{downs2022google} and PartObjaverse-Tiny~\cite{yang2024sampart3d}, spanning single-part geometries and multi-part assemblies. 
Each model has 3 editing prompts, giving 300 editing cases. Each prompt provides 2D renderings of the source, a 2D mask of the edit region, an edited image that serves as the target condition $c^{\mathrm{tar}}$, and a human-annotated 3D mask of the edit region. We use this set because the local metrics need explicit 3D region annotations: the 3D mask only partitions sampled surface points into edited and preserved sets for evaluation.

\paragraph{Implementation Details.}
We use an Euler ODE solver configured for $T=50$ time steps. We initiate the edit trajectory at step $n_{\mathrm{max}}=41$ (skipping the first $9$ integration steps) and stop at step $n_{\mathrm{min}} = 1$, following AnchorFlow~\cite{zhou2025anchorflow}.
Intrinsic spatial localization extracts cross-attention from transformer blocks $\mathcal{B} = \{4,5\}$, with mask parameters $\alpha=0.2$ and $\gamma=1.0$. The guidance phase activates at the discrete step boundary $s^{\mathrm{mask}}=5$ and ramps over $s^{\mathrm{smooth}}=15$ steps with a reference weight $w=0.3$.
The reference latent $z^{\mathrm{ref}}$ is generated with 30 denoising steps. 
The target and source guidance scales for the edit trajectory are 7.5 and 3.5, respectively.
Sweeps over $\alpha$, $\gamma$, $w$, $s^{\mathrm{mask}}$, and the number of reference steps are reported in Appendix~\ref{sec:supp:sens}.

\paragraph{Baselines.}
We compare \ours{} against five representative 3D editing frameworks. \textit{Direct Editing}~\cite{hunyuan3d2025hunyuan3d} generates a new mesh from scratch, conditioned on the target image. \textit{Inversion}~\cite{jiao2025uniedit} performs deterministic latent inversion prior to executing forward generation. For native flow-based editing, we evaluate \textit{Inversion-free}~\cite{kulikov2025flowedit}, which integrates unconstrained velocity differences, and \textit{AnchorFlow}~\cite{zhou2025anchorflow}, which stabilizes the trajectory via latent-anchor alignment. We also include \textit{Nano3D}~\cite{ye2025nano3d} as a contemporary training-free baseline. Notably, \ours{} needs no manual spatial annotations or segmentation masks during editing.

\paragraph{Metrics.}
Four geometric metrics are computed on output  meshes~\cite{fan2017point}. 
No metric uses a ground-truth edited mesh.
To quantify structural integrity on the preserved points, we measure the Chamfer Distance (CD$_{\mathrm{preserve}}\!\downarrow$) and Normal Consistency (NC$_{\mathrm{preserve}}\!\uparrow$)~\cite{gkioxari2019mesh}.
To measure the size of the change, we compute the same distance on the edited points (CD$_{\mathrm{edit}}\!\uparrow$); a higher value indicates a larger change.
On the preserved points, we also compute the F-score$\!\uparrow$~\cite{Knapitsch2017} of the edited mesh against the source mesh: the harmonic mean of precision and recall at a distance threshold of $\tau=0.01$.
We measure semantic alignment using the cosine similarity between target embeddings and six uniformly sampled gray renders of the edited mesh (CLIP$_{\mathrm{img}}\!\uparrow$ and CLIP$_{\mathrm{txt}}\!\uparrow$)~\cite{radford2021learning}.

\subsection{Research Questions}
\label{sec:exp:rqs}
We organize our empirical study around four research questions:
\begin{itemize}
  \item \textbf{RQ1 (State-of-the-Art Comparison):} Does the phase-scheduled integration of Feature-Divergence Localization (FDL) and Reference Generation Guidance (RGG) yield stronger semantic modification and better source-identity preservation than global baselines?
  \item \textbf{RQ2 (Phase Scheduling Necessity):} Does the discrete temporal boundary $s^{\mathrm{mask}}$ between the two phases affect edit strength and semantic alignment? The ``Concurrent'' ablation removes this boundary to test this.
  \item \textbf{RQ3 (Transition Dynamics):} How do mask softness and temporal weighting shape the resulting geometry? The ``Binary Mask'' and ``Instantaneous Activation'' ablations isolate each of these two factors.
  \item \textbf{RQ4 (Component Independence):} Are FDL and RGG jointly necessary to balance semantic editing against geometric preservation, or does either suffice on its own? The ``No Mask'' and ``No Latent Reference'' ablations test each component in turn.
\end{itemize}

\begin{figure}[t!]
  \centering
  \includegraphics[width=\linewidth]{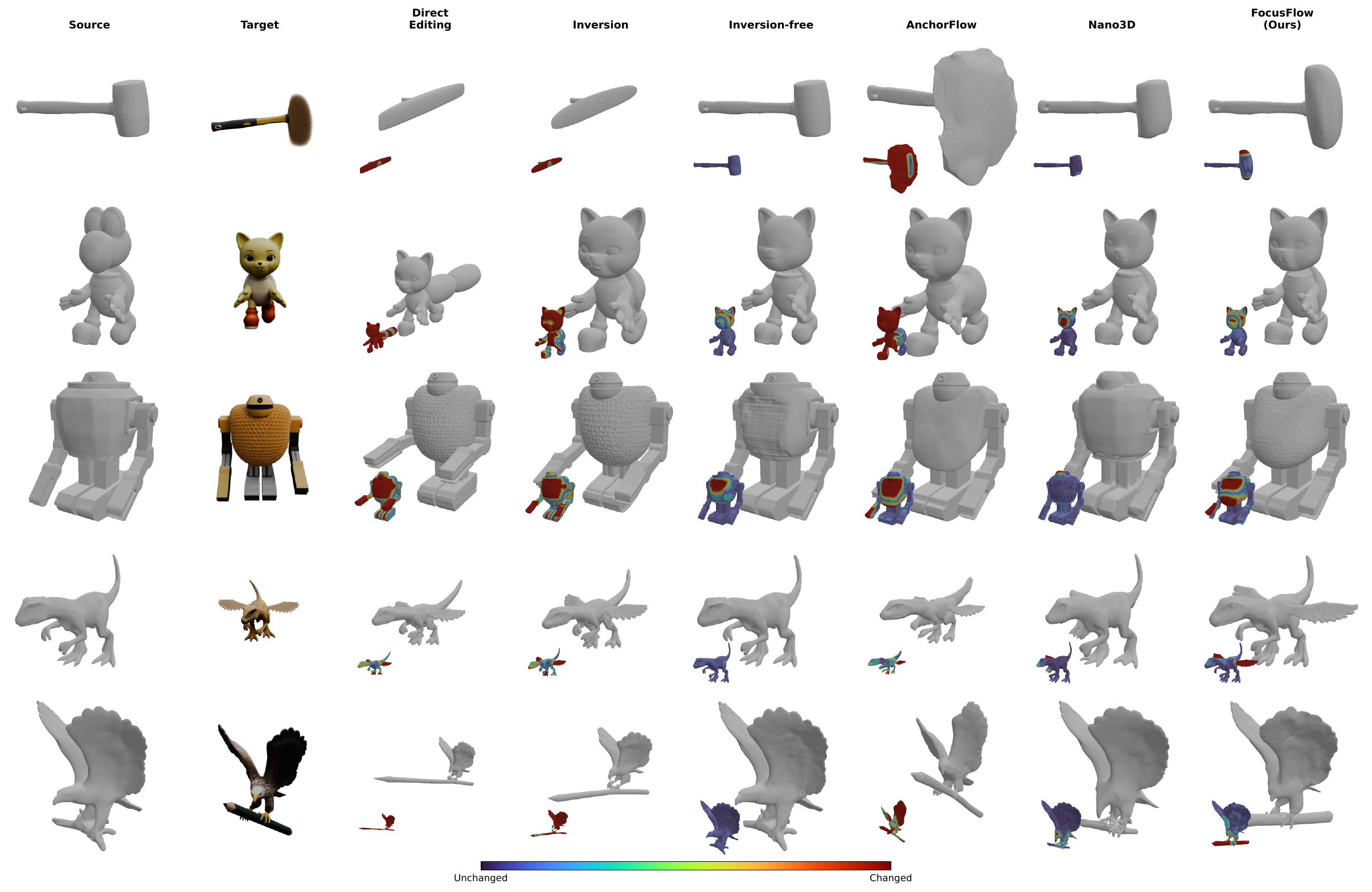}
  \vspace{-0.3cm}
  \caption{\textbf{Qualitative comparison of 3D editing baselines.}
  The inset under each result shows the per-point deviation of the output from the source mesh (blue: unchanged, red: changed), so a well-localized edit appears blue outside the edit region.
  While prior methods either alter the source geometry or fail to fully realize the target modification, \ours{} realizes the edit while leaving unedited regions largely intact.}
  \label{fig:qualitative}
  \vskip -0.2in
\end{figure}

\subsection{Main Results}
\label{sec:exp:main}

To address \textbf{RQ1}, we evaluate \ours{} against external baselines on the GSO and PartObjaverse-Tiny sets in Table~\ref{tab:main}. 
\ours{} has the best value on all four geometric metrics:
$\mathrm{CD}_{\mathrm{preserve}}$ of 3.11, $\mathrm{CD}_{\mathrm{edit}}$ of 1.61,
$\mathrm{NC}_{\mathrm{preserve}}$ of 58.20, and F-score of 3.05. 
It ranks second to Inversion on both CLIP metrics. These suggest that \ours{} can reduce the trade-off between structural preservation and semantic modification.

Table~\ref{tab:main} also reflects different failure modes among the baselines. Methods that regenerate the latent embeddings from the target condition~\cite{hunyuan3d2025hunyuan3d, jiao2025uniedit} reach high semantic alignment but distort the unedited regions, yielding the worst $\mathrm{CD}_{\mathrm{preserve}}$ scores (4.66 and 3.94).
Conversely, the unconstrained flow method~\cite{kulikov2025flowedit} stays close to the source, but has the lowest $\mathrm{CLIP}_{\mathrm{txt}}$ (21.72), consistent with trajectory under-commitment. 
Nano3D~\cite{ye2025nano3d} ranks third on $\mathrm{CD}_{\mathrm{preserve}}$ and $\mathrm{NC}_{\mathrm{preserve}}$, with lower semantic alignment ($\mathrm{CLIP}_{\mathrm{txt}}$ of 22.08).
By combining intrinsic spatial localization and reference generation guidance on geometry, \ours{} reaches semantic alignment between the two regeneration methods with less spatial bleed, so Table~\ref{tab:main} supports \textbf{RQ1} under this protocol.

\begin{table}[hbp]
  \centering
  \caption{Component ablations on the combined GSO and PartObjaverse-Tiny set. 
  }
  \label{tab:abl-component}
  \small
  \scalebox{0.9}{
  \begin{tabular}{lcccccc}
    \toprule
    Variant
      & $\mathrm{CD}_{\mathrm{preserve}}\!\downarrow$
      & $\mathrm{CD}_{\mathrm{edit}}\!\uparrow$
      & $\mathrm{NC}_{\mathrm{preserve}}\!\uparrow$
      & $\mathrm{F\text{-}score}\!\uparrow$
      & $\mathrm{CLIP}_{\mathrm{img}}\!\uparrow$
      & $\mathrm{CLIP}_{\mathrm{txt}}\!\uparrow$ \\
    \midrule
No Mask           & 3.23 & 1.55          & 56.87 & 2.93 & 63.40 & 23.60 \\
    Binary Mask       & 3.38 & 1.57          & 57.18 & 2.86 & 63.35 & 23.54 \\
    Concurrent        & 3.13 & 1.51          & 57.32 & 2.72 & 62.79 & 23.11 \\
    No Latent Ref     & 3.32 & 1.60          & 57.19 & 2.69 & 63.26 & 23.49 \\
    Instant. Act.     & 3.24 & \textbf{1.77} & 57.02 & 2.79 & 63.56 & 23.54 \\
    \midrule
    \textbf{FocusFlow (Ours)}
                      & \textbf{3.11} & 1.61 & \textbf{58.20}
                      & \textbf{3.05} & \textbf{63.85} & \textbf{23.91} \\
    \bottomrule
  \end{tabular}
  }
  \vskip -0.15in
\end{table}

\subsection{Qualitative Evaluation}
\label{sec:exp:qual}

Figure~\ref{fig:qualitative} illustrates, on part addition and replacement, the behaviors quantified in Section~\ref{sec:exp:main}. The baseline methods manifest the failure modes described in Section~\ref{sec:prelim}. Regenerative frameworks lack intrinsic spatial localization, distorting unedited regions and degrading source identity (rows 1 and 5, cols 3--4). Conversely, unconstrained flow methods often suffer from trajectory under-commitment, stalling before realizing the semantic target (rows 1, 3, 4, and 5, col 5). Even when stabilized by latent anchoring, global velocity updates often cause spatial bleed, modifying unedited background geometries (rows 1, 2, 4, and 5, col 6). Furthermore, methods operating on coupled representations can under-edit (row 1, col 7) or introduce structural artifacts (row 3, col 7).

\ours{} reduces these failures. By enforcing intrinsic spatial localization during coarse structure formation, it gates the velocity field to preserve the source geometry. Subsequently applying time-dependent weighting toward the reference latent mitigates the abrupt change in the velocity field, driving convergence to the target shape. In Figure~\ref{fig:qualitative}, \ours{} confines most of the change to the edit: in the deviation insets, the handle stays blue while the head changes (row 1, col 8), the body stays blue while wings are added (row 4, col 8), and the eagle stays mostly blue while the branch becomes a pencil (row 5, col 8).

\subsection{Ablation Studies}
\label{sec:exp:abl}

\begin{figure}[t!]
  \centering
  \includegraphics[width=1.0\linewidth]{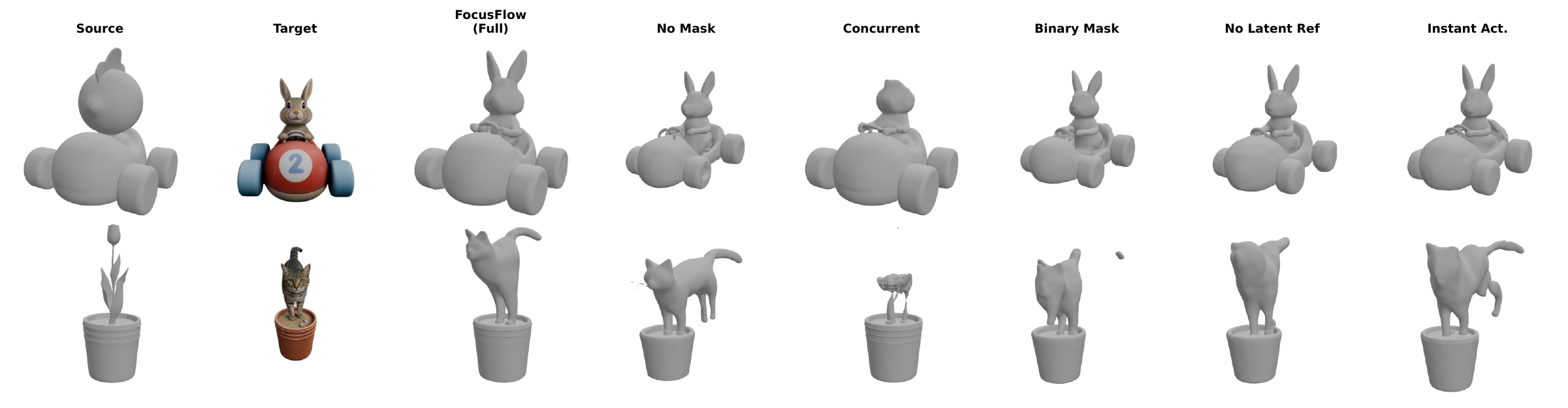} 
  \caption{\textbf{Qualitative ablation results.} Visual comparison of ablated variants against the full framework on two part-replacement cases.}
  \label{fig:ablation}
\end{figure}

To isolate the marginal contributions of the proposed mechanisms and address \textbf{RQ2}, \textbf{RQ3}, and \textbf{RQ4}, we evaluate five ablated variants on the joint benchmark (Table~\ref{tab:abl-component}); Fig.~\ref{fig:ablation} shows two example cases.

\paragraph{Phase Scheduling Necessity (RQ2).}
We evaluate the requirement for the temporal boundary $s^{\mathrm{mask}}$ by activating intrinsic spatial localization and reference generation guidance simultaneously across all integration steps (\textit{Concurrent}). This simultaneous activation keeps the background close to the source but suppresses the intended edit: the chick head stays in row 1, and no cat forms in row 2. It yields the weakest edit magnitude and the poorest semantic alignment in Table~\ref{tab:abl-component} ($\mathrm{CLIP}_{\mathrm{txt}}$ of $23.11$), which supports separating the mechanisms into disjoint phases. 
The late pull also shows no clear harm to preservation: the full model has the lowest $\mathrm{CD}_{\mathrm{preserve}}$ in Table~\ref{tab:abl-component} (3.11, against 3.32 for \textit{No Latent Ref}, which applies no pull).

\paragraph{Transition Dynamics (RQ3).}
We assess the impact of transition functions via the \textit{Binary Mask} and \textit{Instant.\ Act.} ablations.
Replacing the soft mask with a strict binary threshold (\textit{Binary Mask}) gives the highest $\mathrm{CD}_{\mathrm{preserve}}$ in Table~\ref{tab:abl-component} and, in Fig.~\ref{fig:ablation}, leaves a floating fragment (row 2).
Furthermore, substituting the time-dependent weighting with an instantaneous activation creates an abrupt shift in the velocity field. 
While this sudden update gives the highest $\mathrm{CD}_{\mathrm{edit}}$, it can induce topological fracturing, the mesh tearing into disconnected fragments, as with the detached leg in row 2, and it lowers $\mathrm{NC}_{\mathrm{preserve}}$ and the F-score against the full model. These results suggest that the smooth ramp helps avoid such artifacts.

\paragraph{Component Independence (RQ4).}
In Table~\ref{tab:abl-component}, removing either mechanism worsens every metric against the full model, which suggests that both contribute to the balance. Disabling FDL (\textit{No Mask}) causes spatial bleed into background geometry, shrinking the pot (row 2), and raises $\mathrm{CD}_{\mathrm{preserve}}$ to 3.23.
Omitting RGG (\textit{No Latent Ref}) leans toward trajectory under-commitment: integrating velocity differences alone leaves the target shape less fully realized, as with the faceless cat in row 2, and lowers semantic alignment ($\mathrm{CLIP}_{\mathrm{img}}$). It also gives the lowest preservation F-score in the table.

On a single A100, \ours{} takes 29.82\,s per edit, about 12\% more than Inversion (26.56\,s) and about 40\% more than AnchorFlow (21.38\,s).
Appendices~\ref{sec:supp:comp} and~\ref{sec:supp:sens} detail computational cost and hyperparameter sensitivity.
Appendix~\ref{sec:supp:extra} reports the external benchmarks, robustness to a degraded reference, the Jacobian check, and the user study.

\section{Conclusion}
\label{sec:conclusion}

We introduced \ours{}, a segmentation-free framework that reduces the spatial bleed and trajectory under-commitment of inversion-free probability flow editing. By temporally separating intrinsic spatial localization (Feature-Divergence Localization) and terminal target convergence (Reference Generation Guidance), the integration trajectory protects unedited geometry during coarse structure formation and drives endpoint convergence once coarse geometry has formed. 
On the joint GSO and PartObjaverse-Tiny benchmark, the phase-scheduled integration obtains the best preservation scores and the largest edit magnitude among compared methods. This suggests that it can reduce the conventional trade-off between geometric preservation and semantic edit magnitude without requiring manual spatial annotations during editing.
More broadly, decomposing flow trajectories into temporally specialized phases offers a useful design pattern wherever preservation and modification compete. We view this as a step toward phase-aware scheduling for controllable generative editing more generally.

\section{Limitations}
\label{sec:supp:limit}
Despite improving trajectory stability, the method exhibits specific limitations. First, the final mesh depends strictly on the frozen ShapeVAE~\cite{hunyuan3d2025hunyuan3d}. Its compression of high-frequency geometry inevitably degrades fine source details, capping the maximum achievable geometric preservation regardless of latent precision. As a training-free method, however, \ours{} can adopt stronger base generators without retraining. Second, FDL relies on accurate semantic mapping by the frozen DINO encoder~\cite{oquab2023dinov2}. If the patch-wise cosine distance fails to cleanly isolate the modified region, the spatial gate incorrectly bounds the update, causing localized spatial bleed. Third, RGG adds a one-time 30-step reference-generation pass, which raises runtime by about 40\% over AnchorFlow. Finally, relying on a rigidly fixed temporal boundary $s^{\mathrm{mask}}$ may retire the spatial gate too early or too late for certain shapes. We discuss future work and broader impacts in Appendices~\ref{sec:supp:future} and~\ref{sec:supp:impacts}.

\begin{ack}
This work was supported by the Institute of Information \& Communications Technology Planning \& Evaluation (IITP) grant funded by the Korea government (MSIT) (No. RS-2026-25584577, KAIST Deep Tech Bridge Center).
\end{ack}

\newpage








{
\small
\bibliographystyle{plain}
\bibliography{bibliography}
}


\appendix

\newpage

\section{Algorithm}
\label{sec:supp:algo}

Algorithm~\ref{alg:focusflow} details the precise execution. By sequencing the integration into the localization phase ($i < s^{\mathrm{mask}}$) and the guidance phase ($i \ge s^{\mathrm{mask}}$), the spatial gate and the endpoint pull never act on the same step. In the guidance phase, the update subtracts the reference gradient to drive convergence to the target geometry.

\begin{algorithm}[h]
\caption{\ours{} phase-scheduled integration.}
\label{alg:focusflow}
\KwIn{Source mesh, source condition $c^{\mathrm{src}}$, target condition $c^{\mathrm{tar}}$,
      discrete integration steps $T$, edit range $n_{\mathrm{max}}, n_{\mathrm{min}}$,
      discrete phase boundary $s^{\mathrm{mask}}$,
      discrete smoothing duration $s^{\mathrm{smooth}}$,
      preservation floor $\alpha$, sharpness exponent $\gamma$,
      reference weight $w$.}
\KwOut{Edited clean latent $z_1^{\mathrm{edit}}$.}
\BlankLine
\tcp{Initialization}
$\phi^{\mathrm{src}}, \phi^{\mathrm{tar}} \gets \mathcal{E}_{\mathrm{DINO}}(c^{\mathrm{src}}), \mathcal{E}_{\mathrm{DINO}}(c^{\mathrm{tar}})$\;
Define active edit region $K_{\mathrm{changed}}$ where cosine distance $D_p > \mu(D) + \sigma(D)$ \tcp*{Eq.~\ref{eq:dino}}
$z^{\mathrm{ref}} \gets \mathrm{Generator}(c^{\mathrm{tar}})$ \tcp*{Reference latent}
$z_1^{\mathrm{src}}\gets$ ShapeVAE encoding of source mesh\;
$n_{\mathrm{edit}} \gets n_{\mathrm{max}} - n_{\mathrm{min}}$ \tcp*{Number of edit steps}
$z_{t_0}^{\mathrm{edit}} \gets z_1^{\mathrm{src}}$\;
\BlankLine
\For{$i = 0, \dots, n_{\mathrm{edit}} - 1$}{
  $t_i \gets \mathrm{TimeSchedule}(T - n_{\mathrm{max}} + i)$ \tcp*{Skips the first $T - n_{\mathrm{max}}$ steps}

  Sample $\epsilon \sim \mathcal{N}(0, I)$; $z_{t_i}^{\mathrm{src}} \gets t_i z_1^{\mathrm{src}} + (1-t_i)\epsilon$\;
  $z_{t_i}^{\mathrm{tar}} \gets z_{t_i}^{\mathrm{edit}} + z_{t_i}^{\mathrm{src}} - z_1^{\mathrm{src}}$\;
  $v_{t_i}^{\mathrm{edit}} \gets v_\theta(z_{t_i}^{\mathrm{tar}}, t_i, c^{\mathrm{tar}}) - v_\theta(z_{t_i}^{\mathrm{src}}, t_i, c^{\mathrm{src}})$ \;
  
  Compute AnchorFlow alignment gradient $\nabla \mathcal{L}_{\mathrm{align}}$\;
  
  \uIf{$i < s^{\mathrm{mask}}$}{
    \tcp{Localization Phase}
    Extract multi-block cross-attention scores $A_{t_i}^{\mathrm{src}}, A_{t_i}^{\mathrm{tar}}$\;
    Compute $\mathrm{score}_n$ over $K_{\mathrm{changed}}$ \tcp*{Eq.~\ref{eq:attn_score}}
    
    Form soft mask $m$ parameterized by $(\alpha, \gamma)$ \tcp*{Eq.~\ref{eq:mask}}
    $\Delta z_{t_i} \gets \big(v_{t_i}^{\mathrm{edit}} - \nabla \mathcal{L}_{\mathrm{align}}\big) \odot m \, \mathrm{d}t_i$ \tcp*{Eq.~\ref{eq:gated}}
  }
  \Else{
    \tcp{Guidance Phase}
    $\hat{z}_1^{\mathrm{edit}} \gets z_{t_i}^{\mathrm{edit}} + (1 - t_i)v_{t_i}^{\mathrm{edit}}$ \tcp*{Clean-latent forecast}
    $r(i) \gets \min\Big(1, \max\big(0, \frac{i - s^{\mathrm{mask}}}{s^{\mathrm{smooth}}}\big)\Big)$ \tcp*{Trajectory smoothing via step index}
    $\nabla \mathcal{L}_{\mathrm{ref}} \gets (2 - t_i) (\hat{z}_1^{\mathrm{edit}} - z^{\mathrm{ref}})$ \tcp*{Eq.~\ref{eq:lref-approx}}
    $\Delta z_{t_i} \gets \big(v_{t_i}^{\mathrm{edit}} - \nabla \mathcal{L}_{\mathrm{align}} - w\ r(i)\nabla \mathcal{L}_{\mathrm{ref}}\big) \mathrm{d}t_i$ \tcp*{Eq.~\ref{eq:guided}}
  }
  $z_{t_{i+1}}^{\mathrm{edit}} \gets z_{t_i}^{\mathrm{edit}} + \Delta z_{t_i}$\;
}
\Return $z_1^{\mathrm{edit}}$\;
\end{algorithm}

\section{Hyperparameter Sensitivity}
\label{sec:supp:sens}

We evaluate sensitivity along five axes, varying one hyperparameter at a time while holding the others at the defaults defined in Sec.~\ref{sec:exp:setup} ($\alpha = 0.2$, $\gamma = 1.0$, $w = 0.3$, $s^{\mathrm{mask}} = 5$, $s^{\mathrm{smooth}} = 15$, and 30 denoising steps for the reference latent $z^{\mathrm{ref}}$). All evaluations are performed on the joint GSO and PartObjaverse-Tiny benchmark.

\paragraph{Preservation floor $\alpha$.}
The floor $\alpha$ is the minimum fraction of the update that every token receives during the localization phase. Table~\ref{tab:supp:alpha} sweeps $\alpha \in \{0.0, 0.2, 0.4, 0.6\}$. The default $\alpha = 0.2$ achieves the optimal performance across all metrics. Setting $\alpha = 0$ gives low-score tokens almost no update, which degrades $\mathrm{CD}_{\mathrm{preserve}}$ to $3.44$. Larger floors ($\alpha \ge 0.4$) allow the edit signal to flood into preservation tokens, monotonically worsening $\mathrm{CD}_{\mathrm{preserve}}$ from $3.51$ at $\alpha = 0.4$ to $3.65$ at $\alpha = 0.6$.

\begin{table}[h]
  \centering
  \caption{Sensitivity to the preservation floor $\alpha$.}
  \label{tab:supp:alpha}
  \small
  \begin{tabular}{ccccccc}
    \toprule
    $\alpha$
      & $\mathrm{CD}_{\mathrm{preserve}}\!\downarrow$
      & $\mathrm{CD}_{\mathrm{edit}}\!\uparrow$
      & $\mathrm{NC}_{\mathrm{preserve}}\!\uparrow$
      & $\mathrm{F\text{-}score}\!\uparrow$
      & $\mathrm{CLIP}_{\mathrm{img}}\!\uparrow$
      & $\mathrm{CLIP}_{\mathrm{txt}}\!\uparrow$ \\
    \midrule
    0.0      & 3.44          & 1.50          & 56.86          & 2.43          & 63.06          & 23.21 \\
    \textbf{0.2} & \textbf{3.11} & \textbf{1.61} & \textbf{58.20} & \textbf{3.05} & \textbf{63.85} & \textbf{23.91} \\
    0.4      & 3.51          & 1.40          & 56.73          & 2.58          & 62.44          & 22.97 \\
    0.6      & 3.65          & 1.45          & 56.74          & 2.57          & 62.62          & 23.00 \\
    \bottomrule
  \end{tabular}
\end{table}

\paragraph{Sharpness exponent $\gamma$.}
The exponent $\gamma$ dictates the transition speed between preservation (near $\alpha$) and edit (near $1$) states. Table~\ref{tab:supp:gamma} reports $\gamma \in \{0.5, 1.0, 2.0\}$. The linear regime ($\gamma = 1.0$) provides the best balance. A flatter mask ($\gamma = 0.5$) inflates $\mathrm{CD}_{\mathrm{edit}}$ to $1.65$ by spreading the edit signal but significantly degrades preservation metrics ($\mathrm{CD}_{\mathrm{preserve}}$ $3.52$, $\mathrm{NC}_{\mathrm{preserve}}$ $56.65$). Sharper masks ($\gamma = 2.0$) approach a binary gate, recovering the degradation patterns discussed in Sec.~\ref{sec:exp:abl}.

\begin{table}[h]
  \centering
  \caption{Sensitivity to the sharpness exponent $\gamma$.}
  \label{tab:supp:gamma}
  \small
  \begin{tabular}{ccccccc}
    \toprule
    $\gamma$
      & $\mathrm{CD}_{\mathrm{preserve}}\!\downarrow$
      & $\mathrm{CD}_{\mathrm{edit}}\!\uparrow$
      & $\mathrm{NC}_{\mathrm{preserve}}\!\uparrow$
      & $\mathrm{F\text{-}score}\!\uparrow$
      & $\mathrm{CLIP}_{\mathrm{img}}\!\uparrow$
      & $\mathrm{CLIP}_{\mathrm{txt}}\!\uparrow$ \\
    \midrule
    0.5      & 3.52          & \textbf{1.65} & 56.65          & 2.56          & 62.32          & 23.01 \\
    \textbf{1.0} & \textbf{3.11} & 1.61          & \textbf{58.20} & \textbf{3.05} & \textbf{63.85} & \textbf{23.91} \\
    2.0      & 3.52          & 1.53          & 56.74          & 2.38          & 62.83          & 23.24 \\
    \bottomrule
  \end{tabular}
\end{table}

\paragraph{Reference weight $w$.}
This weight balances the differential edit velocity $v_t^{\mathrm{edit}}$ against the clean-latent residual. As shown in Table~\ref{tab:supp:weight}, the default $w = 0.3$ is optimal. While metrics exhibit some non-monotonicity at intermediate weights, $\mathrm{CLIP}_{\mathrm{img}}$ falls from $63.85$ at $w = 0.3$ to $60.88$ at $w = 1.5$. This confirms the trajectory collapse predicted in Sec.~\ref{sec:method:rgg}, where the residual dominates the update and reverts the edit toward a direct copy of $z^{\mathrm{ref}}$, erasing the source identity.

\begin{table}[h]
  \centering
  \caption{Sensitivity to the reference weight $w$.}
  \label{tab:supp:weight}
  \small
  \begin{tabular}{ccccccc}
    \toprule
    $w$
      & $\mathrm{CD}_{\mathrm{preserve}}\!\downarrow$
      & $\mathrm{CD}_{\mathrm{edit}}\!\uparrow$
      & $\mathrm{NC}_{\mathrm{preserve}}\!\uparrow$
      & $\mathrm{F\text{-}score}\!\uparrow$
      & $\mathrm{CLIP}_{\mathrm{img}}\!\uparrow$
      & $\mathrm{CLIP}_{\mathrm{txt}}\!\uparrow$ \\
    \midrule
    \textbf{0.3} & \textbf{3.11} & \textbf{1.61} & \textbf{58.20} & \textbf{3.05} & \textbf{63.85} & \textbf{23.91} \\
    0.5      & 3.44          & 1.47          & 56.69          & 2.51          & 62.78          & 23.05 \\
    1.0      & 3.40          & 1.45          & 56.65          & 2.56          & 62.24          & 22.89 \\
    1.5      & 3.52          & 1.54          & 56.65          & 2.51          & 60.88          & 22.29 \\
    \bottomrule
  \end{tabular}
\end{table}

\paragraph{Mask boundary $s^{\mathrm{mask}}$.}
Table~\ref{tab:supp:mask} sweeps the discrete phase boundary. The shortest setting ($s^{\mathrm{mask}} = 5$) yields the best results. Extending the localization phase beyond this point ties the spatial gate to fine detail rather than coarse layout, causing $\mathrm{CD}_{\mathrm{preserve}}$ to rise from $3.11$ to $3.41$ as $s^{\mathrm{mask}}$ increases to $10$. Similarly, the F-score drops from $3.05$ to $2.39$ as the guidance pull is delayed until step $15$.

\begin{table}[h]
  \centering
  \caption{Sensitivity to the discrete mask boundary $s^{\mathrm{mask}}$.}
  \label{tab:supp:mask}
  \small
  \begin{tabular}{ccccccc}
    \toprule
    $s^{\mathrm{mask}}$
      & $\mathrm{CD}_{\mathrm{preserve}}\!\downarrow$
      & $\mathrm{CD}_{\mathrm{edit}}\!\uparrow$
      & $\mathrm{NC}_{\mathrm{preserve}}\!\uparrow$
      & $\mathrm{F\text{-}score}\!\uparrow$
      & $\mathrm{CLIP}_{\mathrm{img}}\!\uparrow$
      & $\mathrm{CLIP}_{\mathrm{txt}}\!\uparrow$ \\
    \midrule
    \textbf{5}   & \textbf{3.11} & \textbf{1.61} & \textbf{58.20} & \textbf{3.05} & \textbf{63.85} & \textbf{23.91} \\
    10      & 3.41          & 1.55          & 56.95          & 2.52          & 62.55          & 23.00 \\
    15      & 3.34          & 1.51          & 56.80          & 2.39          & 62.49          & 22.74 \\
    \bottomrule
  \end{tabular}
\end{table}

\paragraph{Denoising steps for $z^{\mathrm{ref}}$.} 
We vary the number of steps used to precompute the reference latent $z^{\mathrm{ref}}$ while holding the editing budget fixed. Table~\ref{tab:supp:gen} reports results for 20, 30, and 50 steps. Thirty steps provide the optimal balance; reducing to 20 steps leaves $z^{\mathrm{ref}}$ under-converged, weakening the guidance pull and degrading $\mathrm{CD}_{\mathrm{preserve}}$ to $3.41$. Increasing to 50 steps yields marginal gains in semantic alignment but degrades structural preservation, suggesting diminishing returns relative to the overhead.

\begin{table}[h]
  \centering
  \caption{Sensitivity to the number of denoising steps for $z^{\mathrm{ref}}$.}
  \label{tab:supp:gen}
  \small
  \begin{tabular}{ccccccc}
    \toprule
    Steps
      & $\mathrm{CD}_{\mathrm{preserve}}\!\downarrow$
      & $\mathrm{CD}_{\mathrm{edit}}\!\uparrow$
      & $\mathrm{NC}_{\mathrm{preserve}}\!\uparrow$
      & $\mathrm{F\text{-}score}\!\uparrow$
      & $\mathrm{CLIP}_{\mathrm{img}}\!\uparrow$
      & $\mathrm{CLIP}_{\mathrm{txt}}\!\uparrow$ \\
    \midrule
    20      & 3.41          & 1.51          & 56.71          & 2.54          & 62.64          & 23.14 \\
    \textbf{30} & \textbf{3.11} & 1.61          & \textbf{58.20} & 3.05          & 63.85          & 23.91 \\
    50      & 3.14          & \textbf{1.66} & 58.16          & \textbf{3.07} & \textbf{64.19} & \textbf{23.95} \\
    \bottomrule
  \end{tabular}
\end{table}

\section{Implementation Details of Baselines}
\label{sec:supp:baseline_details}

To ensure a rigorous and reproducible comparison, we provide the specific configurations for all external baseline methods. All experiments are conducted using a fixed random seed of 42.

\paragraph{Direct Editing.} This baseline generates a new mesh from scratch with the Hunyuan3D-2.1 flow-based generator~\cite{hunyuan3d2025hunyuan3d}, conditioned on the target image. We utilize a target guidance scale of 5.0 and an integration schedule of 50 denoising steps.

\paragraph{Editing-by-Inversion.} Following the UniEdit-Flow framework~\cite{jiao2025uniedit}, we perform conditional inversion using the source image (guidance scale 1.0) for 50 steps. The subsequent forward generation uses the target image condition with a guidance scale of 5.0, resulting in a total of 100 function evaluations (NFEs).

\paragraph{Inversion-free.} We implement the velocity difference integration based on the FlowEdit formulation~\cite{kulikov2025flowedit} on top of the Hunyuan3D backbone. To ensure a fair comparison with our framework, we use the same schedule as \ours{} ($T = 50$, $n_{\mathrm{max}} = 41$, $n_{\mathrm{min}} = 1$). We set the target and source guidance scales to 7.5 and 3.5, respectively, with $n_{\mathrm{avg}}=1$ noise sample per step.

\paragraph{AnchorFlow.} We adopt the identical temporal integration schedule ($n_{\mathrm{max}}=41$) and guidance scales (7.5/3.5) as the Inversion-free implementation. We incorporate the official alignment gradient parameters to stabilize the trajectory through latent anchoring.

\paragraph{Nano3D.} Unlike the other baselines, Nano3D~\cite{ye2025nano3d} is built on TRELLIS~\cite{xiang2025structured}. We follow the official implementation for the sparse structure encoder. Assets are processed using the authors' provided voxelization and rendering utilities to ensure structural fidelity.

\section{Computational Efficiency}
\label{sec:supp:comp}

\begin{table}[hbp]
  \centering
  \caption{Computational cost comparison. 
  }
  \label{tab:supp:compute}
  \small
  \begin{tabular}{lcc}
    \toprule
    Method & Runtime (s) & VRAM (GB) \\
    \midrule
    Inversion~\cite{jiao2025uniedit} & $26.56 \pm 0.10$ & $11.12$ \\
    AnchorFlow~\cite{zhou2025anchorflow} & $21.38 \pm 0.10$ & $12.50$ \\
    Nano3D~\cite{ye2025nano3d} & $20.68 \pm 0.68$ & $10.53$ \\
    \ours{} & $29.82 \pm 1.73$ & $13.76$ \\
    \bottomrule
  \end{tabular}
\end{table}

We evaluate the computational efficiency of \ours{} against the baseline editing frameworks. Table~\ref{tab:supp:compute} reports the average of 10 inference runtimes and VRAM usage on a single NVIDIA A100 GPU. \ours{} requires a one-time forward pass (30 steps) to generate the reference latent $z^{\mathrm{ref}}$, followed by the 40-step editing integration. This structure avoids the per-instance optimization overhead typical of score distillation sampling (SDS)-based methods~\cite{poole2022dreamfusion} while maintaining competitive runtime against deterministic inversion pipelines.


\section{Additional Evaluation}
\label{sec:supp:extra}

This appendix reports additional experiments that support Section~\ref{sec:exp}.

\subsection{External Benchmarks}
\label{sec:supp:external}

To test \ours{} beyond Edit3D-Bench, we also run it on two external benchmarks, Eval3DEdit~\cite{zhou2025anchorflow} and Nano3D-Edit~\cite{ye2025nano3d}, under their official protocols, with 100 cases each.
We report alignment scores based on CLIP~\cite{radford2021learning}, Uni3D~\cite{zhou2024uni3d}, and DINO~\cite{oquab2023dinov2}. These sets have no 3D region annotations, so we annotated 3D regions manually to compute $\mathrm{CD}_{\mathrm{preserve}}$.

\ours{} has the best value on all six metrics on both sets (Tables~\ref{tab:supp:eval3dedit} and~\ref{tab:supp:nano3dedit}). The CLIP, Uni3D, and DINO scores need no region annotation or reference geometry, so that part of the ranking does not depend on our annotations. The $\mathrm{CD}_{\mathrm{preserve}}$ values here are not comparable with Table~\ref{tab:main}, because the sets and region annotations differ.

On a resolution-normalized $64^3$ voxelization, chosen to stay well-defined on raw scan meshes, all three benchmarks have a median of about one connected component (26-connectivity) and a median surface occupancy of about 0.04, so their shapes are similar in connectivity and spatial extent. We attribute differences in raw face count mainly to scan tessellation.

\begin{table}[h]
  \centering
  \caption{Results on Eval3DEdit (100 cases). We mark the \textbf{best} and \underline{second-best} results.}
  \label{tab:supp:eval3dedit}
  \small
  \scalebox{0.95}{
  \begin{tabular}{lcccccc}
    \toprule
    Method
      & $\mathrm{CLIP}_{\mathrm{img}}\!\uparrow$
      & $\mathrm{CLIP}_{\mathrm{txt}}\!\uparrow$
      & $\mathrm{Uni3D}_{\mathrm{pc}}\!\uparrow$
      & $\mathrm{Uni3D}_{\mathrm{txt}}\!\uparrow$
      & $\mathrm{DINO}_{\mathrm{img}}\!\uparrow$
      & $\mathrm{CD}_{\mathrm{preserve}}\!\downarrow$ \\
    \midrule
    Direct Editing~\cite{hunyuan3d2025hunyuan3d}
      & 56.12 & 43.08 & 27.85 & 17.20 & 46.88 & 0.149 \\
    Inversion~\cite{jiao2025uniedit}
      & 64.64 & 45.83 & 30.03 & 21.53 & 44.53 & 0.082 \\
    Inversion-free~\cite{kulikov2025flowedit}
      & 64.51 & 45.60 & 30.75 & 21.50 & 43.14 & 0.078 \\
    AnchorFlow~\cite{zhou2025anchorflow}
      & 68.26 & \underline{48.00} & 33.75 & 23.39 & 52.28 & 0.124 \\
    Nano3D~\cite{ye2025nano3d}
      & \underline{69.50} & 47.96 & \underline{34.45} & \underline{24.17} & \underline{53.72} & \underline{0.075} \\
    \midrule
    \textbf{\ours{} (Ours)}
      & \textbf{71.73} & \textbf{49.40} & \textbf{35.21} & \textbf{25.02} & \textbf{57.55} & \textbf{0.067} \\
    \bottomrule
  \end{tabular}
  }
\end{table}

\begin{table}[h]
  \centering
  \caption{Results on Nano3D-Edit (100 cases). We mark the \textbf{best} and \underline{second-best} results.}
  \label{tab:supp:nano3dedit}
  \small
  \scalebox{0.95}{
  \begin{tabular}{lcccccc}
    \toprule
    Method
      & $\mathrm{CLIP}_{\mathrm{img}}\!\uparrow$
      & $\mathrm{CLIP}_{\mathrm{txt}}\!\uparrow$
      & $\mathrm{Uni3D}_{\mathrm{pc}}\!\uparrow$
      & $\mathrm{Uni3D}_{\mathrm{txt}}\!\uparrow$
      & $\mathrm{DINO}_{\mathrm{img}}\!\uparrow$
      & $\mathrm{CD}_{\mathrm{preserve}}\!\downarrow$ \\
    \midrule
    Direct Editing~\cite{hunyuan3d2025hunyuan3d}
      & \underline{60.34} & 40.15 & 29.80 & 14.50 & \underline{45.28} & 0.2591 \\
    Inversion~\cite{jiao2025uniedit}
      & 59.18 & \underline{42.78} & 31.72 & 16.02 & 42.93 & 0.1352 \\
    Inversion-free~\cite{kulikov2025flowedit}
      & 59.44 & 41.37 & 31.95 & 16.00 & 41.54 & 0.1287 \\
    AnchorFlow~\cite{zhou2025anchorflow}
      & 59.43 & 42.66 & 31.61 & 16.05 & 43.86 & 0.1265 \\
    Nano3D~\cite{ye2025nano3d}
      & 59.08 & 42.14 & \underline{32.38} & \underline{16.43} & 41.00 & \underline{0.1138} \\
    \midrule
    \textbf{\ours{} (Ours)}
      & \textbf{62.02} & \textbf{43.43} & \textbf{33.56} & \textbf{17.06} & \textbf{48.01} & \textbf{0.0803} \\
    \bottomrule
  \end{tabular}
  }
\end{table}


\subsection{Robustness to a Misaligned or Failed Reference}
\label{sec:supp:robust}

The experiments in this subsection and in Appendix~\ref{sec:supp:jacobian} use a subset of 100 of the 300 editing cases. RGG relies on an independently generated $z^{\mathrm{ref}}$, which can be misaligned with the source or fail outright. To test how \ours{} behaves in these cases at $w=0.3$, we degrade $z^{\mathrm{ref}}$ in three ways: a $60^\circ$ rigid rotation; over several random seeds, the reference whose preserved region deviates most from the source; and a 5-step under-converged reference. Fracturing would show as the mesh splitting into pieces, so we count connected components, and we also track $\mathrm{CD}_{\mathrm{edit}}$ and CLIP similarity to the source and target. In aggregate across all three degradations, the component count stays flat at about 14 and $\mathrm{NC}_{\mathrm{preserve}}$ barely moves, while the edit weakens: $\mathrm{CD}_{\mathrm{edit}}$ drops from 1.45 to 1.25, and the output stays closer to the source than to the target.

We also label each of 100 test cases under a degraded reference as a warp (preservation breaks and new components appear) or a safe under-commit (weaker edit, source-like output). Table~\ref{tab:supp:refrobust} reports about 90\% under-commit and about 10\% warp. A degraded reference therefore usually costs edit strength rather than source integrity, but it warps the source in 8 to 11 of the 100 cases.

Because $z^{\mathrm{ref}}$ is a one-time precompute, a failed reference can also be screened. Keeping the best of $K$ candidates by CLIP similarity to the target, which needs no ground truth, recovers most of the loss by $K=4$ ($\mathrm{CD}_{\mathrm{preserve}}$ from 3.80 to 3.50, F-score from 2.05 to 2.30).

\begin{table}[h]
  \centering
  \caption{Outcomes over 100 test cases under a degraded $z^{\mathrm{ref}}$ at $w=0.3$ (case counts).}
  \label{tab:supp:refrobust}
  \small
  \begin{tabular}{lccc}
    \toprule
    Degraded reference & Under-commit (safe) & Warp (unsafe) & $N$ \\
    \midrule
    5-step $z^{\mathrm{ref}}$ & 90 & 10 & 100 \\
    $60^\circ$-rotated $z^{\mathrm{ref}}$ & 89 & 11 & 100 \\
    Worst-seed $z^{\mathrm{ref}}$ & 92 & 8 & 100 \\
    \bottomrule
  \end{tabular}
\end{table}


\subsection{Validity of the Scalar-Identity Jacobian Approximation}
\label{sec:supp:jacobian}

We measure the error directly. The exact gradient needs one vector-Jacobian product per guidance step, which we compute over 100 sampled edits and compare with the scalar form. Table~\ref{tab:supp:jacobian} reports the cosine similarity (rotation) and norm ratio (stretch), binned by $t$. In these bins, the cosine similarity averages 0.934 and rises from 0.883 to 0.971 as $t \to 1$, and the norm ratio ranges from 1.05 to 1.48. The bins cover only $t \ge 0.80$, the late part of the guidance phase, which starts near $t = 0.29$.

Using the exact gradient changes the final metrics modestly: $\mathrm{CD}_{\mathrm{preserve}}$ from 3.32 to 3.24, $\mathrm{CD}_{\mathrm{edit}}$ from 1.45 to 1.35, F-score from 2.38 to 2.50, and CLIP within 0.2. It also raises the runtime by about 6.4 times in this experiment (28 to 180\,s).

\begin{table}[h]
  \centering
  \caption{Exact gradient $g_{\mathrm{exact}}$ against the scalar form $g_{\mathrm{approx}}$ over 100 sampled edits, binned by flow time $t$.}
  \label{tab:supp:jacobian}
  \small
  \begin{tabular}{lcc}
    \toprule
    $t$ range & Cosine similarity & $\|g_{\mathrm{exact}}\| / \|g_{\mathrm{approx}}\|$ \\
    \midrule
    0.80--0.85 & 0.883 & 1.48 \\
    0.85--0.90 & 0.929 & 1.17 \\
    0.90--0.95 & 0.956 & 1.28 \\
    0.95--1.00 & 0.971 & 1.05 \\
    \midrule
    \textbf{Overall} & \textbf{0.934} & \textbf{1.23} \\
    \bottomrule
  \end{tabular}
\end{table}


\subsection{Sensitivity of the FDL Threshold on Subtle Edits}
\label{app:threshold_subtle}

FDL selects the edit region $\mathcal{K}_{\mathrm{changed}}$ with the adaptive threshold $D_p > \mu(D) + \sigma(D)$. 
To test this threshold on subtle edits, we stratified the 300 cases by DINO-feature divergence and took the bottom tercile, the 100 smallest-shift cases.
The threshold adapts its selection to the edit: it
selects 9.5\% of the patches on subtle edits versus 16.9\% on large edits, and against the 2D masks derived from the ground-truth 3D edit regions it reaches 83.5\% recall on subtle edits and 80.0\% on large edits. 
Subtle edits have small footprints: the annotated region covers around 4\% of the image, and the selected patches cover about 9\%, so the threshold over-covers the region by roughly 5\% of the image.


\subsection{User Study}
\label{sec:supp:study}

Twenty participants evaluated 30 editing pairs sampled from GSO and PartObjaverse-Tiny. Each case was a two-alternative forced choice~\cite{macmillan2004detection} between \ours{} and one baseline, shown with the source render and target prompt from fixed viewpoints, with left/right order randomized. Participants answered three questions per case: prompt alignment (which model better applies the intended semantic change), visual quality (which shows fewer artifacts such as surface collapse or mesh tearing), and shape preservation (which better preserves the unedited regions and source identity).

Table~\ref{tab:supp:study} reports how often participants preferred \ours{}: 75.7\% to 85.8\% across all comparisons and criteria, highest on shape preservation against AnchorFlow (85.0\%) and Nano3D (85.8\%).

\begin{table}[h]
  \centering
  \caption{User study.}
  \label{tab:supp:study}
  \small
  \begin{tabular}{lccc}
    \toprule
    Baseline & Prompt alignment$\uparrow$ & Visual quality$\uparrow$ & Shape preservation$\uparrow$ \\
    \midrule
    vs. Inversion~\cite{jiao2025uniedit} & 77.8\% & 77.2\% & 77.0\% \\
    vs. AnchorFlow~\cite{zhou2025anchorflow} & 75.7\% & 78.7\% & 85.0\% \\
    vs. Nano3D~\cite{ye2025nano3d} & 75.8\% & 84.5\% & 85.8\% \\
    \bottomrule
  \end{tabular}
\end{table}


\section{Extended Qualitative Results}
\label{sec:supp:qualitative}

\begin{figure}[htbp]
  \centering
  \includegraphics[width=\linewidth]{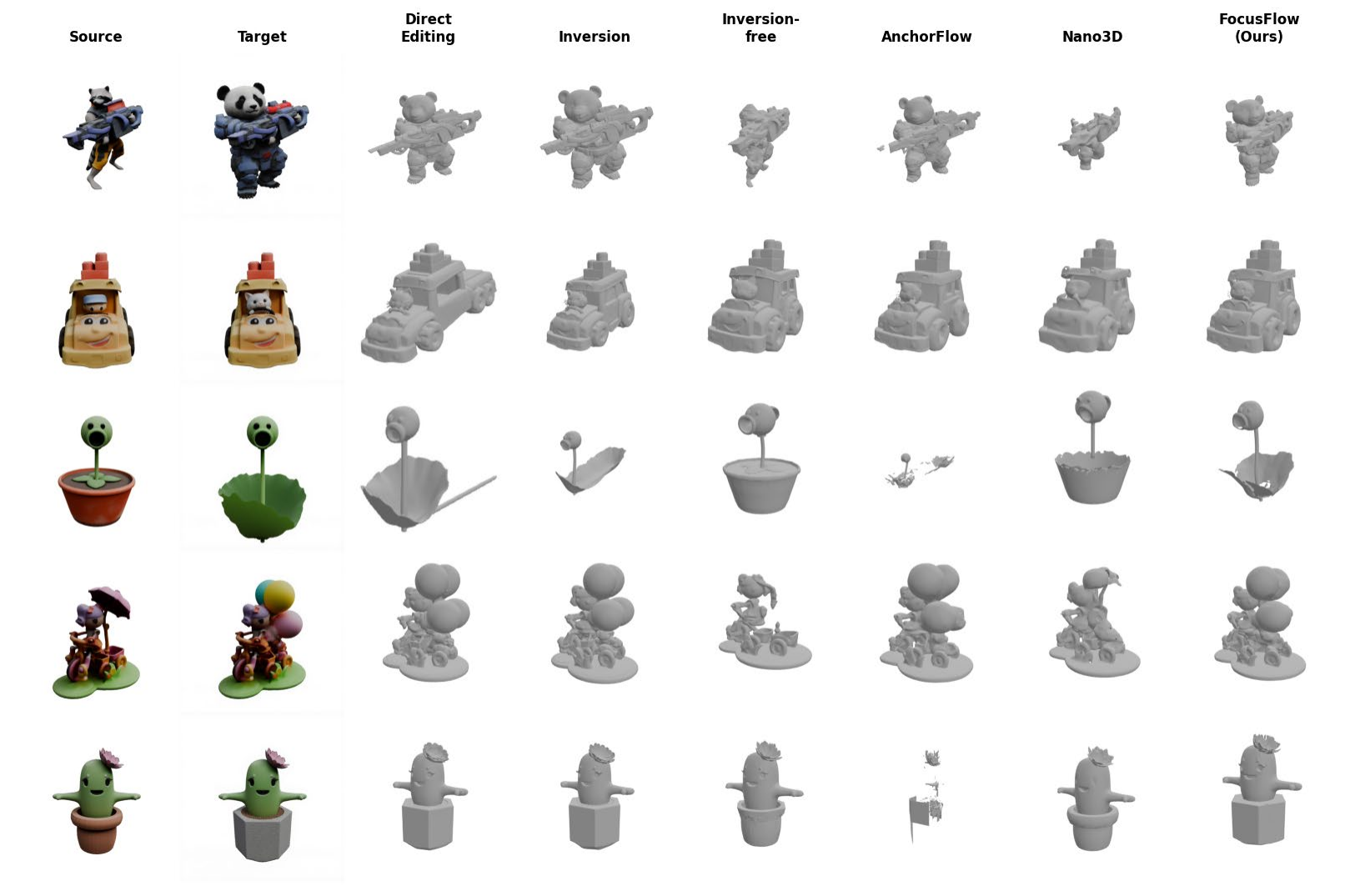} 
  \caption{\textbf{Extended qualitative comparison of 3D editing baselines.} \ours{} realizes the target while leaving most of the unedited geometry unchanged; the baselines show the spatial bleed or trajectory under-commitment.}
  \label{fig:supp:extended_qual}
\end{figure}

Due to space constraints in the main text, we provide an extended visual comparison of \ours{} against the established baseline methods in Figure~\ref{fig:supp:extended_qual}. 

Consistent with the analysis in Section~\ref{sec:exp:main}, regenerative methods~\cite{jiao2025uniedit} successfully capture the semantic target but frequently distort unedited regions. Conversely, unconstrained flow methods~\cite{kulikov2025flowedit} often suffer from trajectory under-commitment, failing to fully realize the target modifications. Global updates with latent anchoring~\cite{zhou2025anchorflow} improve target realization but suffer from severe spatial bleed into the background geometry. Across these cases, which span part addition, part replacement, and whole-object changes, \ours{} realizes the target while leaving most of the unedited source geometry unchanged.

\section{Future Work}
\label{sec:supp:future}

In future work, we plan to extend \ours{} in three directions. First, FDL and RGG modify only the sampling trajectory, so they could be combined with other training-free methods, such as concurrent tangent-space steering~\cite{lim2026tango}, and carried over to scene-level generation~\cite{koo2026inspace, ngo2026volfill}, provided the generator cross-attends to image features. Editing a captured scene would also require reconstructing it, which feed-forward Gaussian splatting does from multi-view images in a single pass~\cite{park2026ecosplat, bui2026airsplat}. Second, the phase boundary and the balancing coefficient are fixed across edits. A reward learned from a few pairwise human judgments~\cite{luu2026video} could select them per edit, and preference optimization~\cite{tee2026diffusion}, with pseudo-preferences generated without human annotation~\cite{luu2026progress}, could align the base generator itself. Third, phase-scheduled integration is not specific to 3D: audio synthesis~\cite{ton2025taro, pham2024mdsgen} and virtual try-on~\cite{hong2025ita} could adopt it without retraining, changing one sound event or one garment while keeping the rest of the output. Conversely, bootstrapping flow-matching velocities into a one-step map~\cite{nguyen2026fast} could shorten the reference pass of RGG.

\section{Broader Impacts}
\label{sec:supp:impacts}

The proposed framework lowers the technical barrier to fine-grained 3D asset editing, which can benefit creators in gaming, spatial computing, and digital design. However, advancing the seamless, training-free modification of synthetic 3D media inherently introduces provenance risks, as the ability to convincingly alter existing structures without obvious artifacts could be misused for unauthorized asset manipulation or deceptive content generation. To mitigate these risks and safeguard digital authenticity, we strongly recommend that future public deployments of latent 3D editing pipelines integrate robust invisible 3D watermarking techniques or cryptographic metadata protocols to ensure verifiable asset lineage.




\newpage
\section*{NeurIPS Paper Checklist}

\begin{enumerate}

\item {\bf Claims}
    \item[] Question: Do the main claims made in the abstract and introduction accurately reflect the paper's contributions and scope?
    \item[] Answer: \answerYes{} 
    \item[] Justification: They can support the main claims of the paper.
    \item[] Guidelines:
    \begin{itemize}
        \item The answer \answerNA{} means that the abstract and introduction do not include the claims made in the paper.
        \item The abstract and/or introduction should clearly state the claims made, including the contributions made in the paper and important assumptions and limitations. A \answerNo{} or \answerNA{} answer to this question will not be perceived well by the reviewers. 
        \item The claims made should match theoretical and experimental results, and reflect how much the results can be expected to generalize to other settings. 
        \item It is fine to include aspirational goals as motivation as long as it is clear that these goals are not attained by the paper. 
    \end{itemize}

\item {\bf Limitations}
    \item[] Question: Does the paper discuss the limitations of the work performed by the authors?
    \item[] Answer: \answerYes{} 
    \item[] Justification: We discuss the limitations in Section \ref{sec:supp:limit}.
    \item[] Guidelines:
    \begin{itemize}
        \item The answer \answerNA{} means that the paper has no limitation while the answer \answerNo{} means that the paper has limitations, but those are not discussed in the paper. 
        \item The authors are encouraged to create a separate ``Limitations'' section in their paper.
        \item The paper should point out any strong assumptions and how robust the results are to violations of these assumptions (e.g., independence assumptions, noiseless settings, model well-specification, asymptotic approximations only holding locally). The authors should reflect on how these assumptions might be violated in practice and what the implications would be.
        \item The authors should reflect on the scope of the claims made, e.g., if the approach was only tested on a few datasets or with a few runs. In general, empirical results often depend on implicit assumptions, which should be articulated.
        \item The authors should reflect on the factors that influence the performance of the approach. For example, a facial recognition algorithm may perform poorly when image resolution is low or images are taken in low lighting. Or a speech-to-text system might not be used reliably to provide closed captions for online lectures because it fails to handle technical jargon.
        \item The authors should discuss the computational efficiency of the proposed algorithms and how they scale with dataset size.
        \item If applicable, the authors should discuss possible limitations of their approach to address problems of privacy and fairness.
        \item While the authors might fear that complete honesty about limitations might be used by reviewers as grounds for rejection, a worse outcome might be that reviewers discover limitations that aren't acknowledged in the paper. The authors should use their best judgment and recognize that individual actions in favor of transparency play an important role in developing norms that preserve the integrity of the community. Reviewers will be specifically instructed to not penalize honesty concerning limitations.
    \end{itemize}

\item {\bf Theory assumptions and proofs}
    \item[] Question: For each theoretical result, does the paper provide the full set of assumptions and a complete (and correct) proof?
    \item[] Answer: \answerNA{} 
    \item[] Justification: The paper does not include theoretical results.
    \item[] Guidelines:
    \begin{itemize}
        \item The answer \answerNA{} means that the paper does not include theoretical results. 
        \item All the theorems, formulas, and proofs in the paper should be numbered and cross-referenced.
        \item All assumptions should be clearly stated or referenced in the statement of any theorems.
        \item The proofs can either appear in the main paper or the supplemental material, but if they appear in the supplemental material, the authors are encouraged to provide a short proof sketch to provide intuition. 
        \item Inversely, any informal proof provided in the core of the paper should be complemented by formal proofs provided in appendix or supplemental material.
        \item Theorems and Lemmas that the proof relies upon should be properly referenced. 
    \end{itemize}

    \item {\bf Experimental result reproducibility}
    \item[] Question: Does the paper fully disclose all the information needed to reproduce the main experimental results of the paper to the extent that it affects the main claims and/or conclusions of the paper (regardless of whether the code and data are provided or not)?
    \item[] Answer: \answerYes{} 
    \item[] Justification: We provide exhaustive implementation details, including solver configurations and exact hyperparameter settings in Section \ref{sec:exp:setup}, as well as a complete step-by-step pseudocode in Appendix \ref{sec:supp:algo} to facilitate full reproducibility.
    \item[] Guidelines:
    \begin{itemize}
        \item The answer \answerNA{} means that the paper does not include experiments.
        \item If the paper includes experiments, a \answerNo{} answer to this question will not be perceived well by the reviewers: Making the paper reproducible is important, regardless of whether the code and data are provided or not.
        \item If the contribution is a dataset and\slash or model, the authors should describe the steps taken to make their results reproducible or verifiable. 
        \item Depending on the contribution, reproducibility can be accomplished in various ways. For example, if the contribution is a novel architecture, describing the architecture fully might suffice, or if the contribution is a specific model and empirical evaluation, it may be necessary to either make it possible for others to replicate the model with the same dataset, or provide access to the model. In general. releasing code and data is often one good way to accomplish this, but reproducibility can also be provided via detailed instructions for how to replicate the results, access to a hosted model (e.g., in the case of a large language model), releasing of a model checkpoint, or other means that are appropriate to the research performed.
        \item While NeurIPS does not require releasing code, the conference does require all submissions to provide some reasonable avenue for reproducibility, which may depend on the nature of the contribution. For example
        \begin{enumerate}
            \item If the contribution is primarily a new algorithm, the paper should make it clear how to reproduce that algorithm.
            \item If the contribution is primarily a new model architecture, the paper should describe the architecture clearly and fully.
            \item If the contribution is a new model (e.g., a large language model), then there should either be a way to access this model for reproducing the results or a way to reproduce the model (e.g., with an open-source dataset or instructions for how to construct the dataset).
            \item We recognize that reproducibility may be tricky in some cases, in which case authors are welcome to describe the particular way they provide for reproducibility. In the case of closed-source models, it may be that access to the model is limited in some way (e.g., to registered users), but it should be possible for other researchers to have some path to reproducing or verifying the results.
        \end{enumerate}
    \end{itemize}

\item {\bf Open access to data and code}
    \item[] Question: Does the paper provide open access to the data and code, with sufficient instructions to faithfully reproduce the main experimental results, as described in supplemental material?
    \item[] Answer: \answerNo{} 
    \item[] Justification: Code is not included with this version of the paper. The method is specified by the pseudo-code in Appendix~\ref{sec:supp:algo}, the solver settings and hyperparameters in Section~\ref{sec:exp:setup} and Appendix~\ref{sec:supp:sens}, and the baseline configurations in Appendix~\ref{sec:supp:baseline_details}.
    \item[] Guidelines:
    \begin{itemize}
        \item The answer \answerNA{} means that paper does not include experiments requiring code.
        \item Please see the NeurIPS code and data submission guidelines (\url{https://neurips.cc/public/guides/CodeSubmissionPolicy}) for more details.
        \item While we encourage the release of code and data, we understand that this might not be possible, so \answerNo{} is an acceptable answer. Papers cannot be rejected simply for not including code, unless this is central to the contribution (e.g., for a new open-source benchmark).
        \item The instructions should contain the exact command and environment needed to run to reproduce the results. See the NeurIPS code and data submission guidelines (\url{https://neurips.cc/public/guides/CodeSubmissionPolicy}) for more details.
        \item The authors should provide instructions on data access and preparation, including how to access the raw data, preprocessed data, intermediate data, and generated data, etc.
        \item The authors should provide scripts to reproduce all experimental results for the new proposed method and baselines. If only a subset of experiments are reproducible, they should state which ones are omitted from the script and why.
        \item At submission time, to preserve anonymity, the authors should release anonymized versions (if applicable).
        \item Providing as much information as possible in supplemental material (appended to the paper) is recommended, but including URLs to data and code is permitted.
    \end{itemize}

\item {\bf Experimental setting/details}
    \item[] Question: Does the paper specify all the training and test details (e.g., data splits, hyperparameters, how they were chosen, type of optimizer) necessary to understand the results?
    \item[] Answer: \answerYes{} 
    \item[] Justification: Section~\ref{sec:exp:setup} gives the solver configuration and all hyperparameters, Appendix~\ref{sec:supp:sens} their sweeps, Appendix~\ref{sec:supp:baseline_details} the exact baseline configurations, and Appendix~\ref{sec:supp:algo} step-by-step pseudo-code.
    \item[] Guidelines:
    \begin{itemize}
        \item The answer \answerNA{} means that the paper does not include experiments.
        \item The experimental setting should be presented in the core of the paper to a level of detail that is necessary to appreciate the results and make sense of them.
        \item The full details can be provided either with the code, in appendix, or as supplemental material.
    \end{itemize}

\item {\bf Experiment statistical significance}
    \item[] Question: Does the paper report error bars suitably and correctly defined or other appropriate information about the statistical significance of the experiments?
    \item[] Answer: \answerNo{} 
    \item[] Justification: Tables~\ref{tab:main},~\ref{tab:abl-component} report means over 300 editing cases without error bars; only runtime has mean ± std (Table~\ref{tab:supp:compute}).
    \item[] Guidelines:
    \begin{itemize}
        \item The answer \answerNA{} means that the paper does not include experiments.
        \item The authors should answer \answerYes{} if the results are accompanied by error bars, confidence intervals, or statistical significance tests, at least for the experiments that support the main claims of the paper.
        \item The factors of variability that the error bars are capturing should be clearly stated (for example, train/test split, initialization, random drawing of some parameter, or overall run with given experimental conditions).
        \item The method for calculating the error bars should be explained (closed form formula, call to a library function, bootstrap, etc.)
        \item The assumptions made should be given (e.g., Normally distributed errors).
        \item It should be clear whether the error bar is the standard deviation or the standard error of the mean.
        \item It is OK to report 1-sigma error bars, but one should state it. The authors should preferably report a 2-sigma error bar than state that they have a 96\% CI, if the hypothesis of Normality of errors is not verified.
        \item For asymmetric distributions, the authors should be careful not to show in tables or figures symmetric error bars that would yield results that are out of range (e.g., negative error rates).
        \item If error bars are reported in tables or plots, the authors should explain in the text how they were calculated and reference the corresponding figures or tables in the text.
    \end{itemize}

\item {\bf Experiments compute resources}
    \item[] Question: For each experiment, does the paper provide sufficient information on the computer resources (type of compute workers, memory, time of execution) needed to reproduce the experiments?
    \item[] Answer: \answerYes{} 
    \item[] Justification: We report the compute resources (NVIDIA A100) in Appendix~\ref{sec:supp:comp}.
    \item[] Guidelines:
    \begin{itemize}
        \item The answer \answerNA{} means that the paper does not include experiments.
        \item The paper should indicate the type of compute workers CPU or GPU, internal cluster, or cloud provider, including relevant memory and storage.
        \item The paper should provide the amount of compute required for each of the individual experimental runs as well as estimate the total compute. 
        \item The paper should disclose whether the full research project required more compute than the experiments reported in the paper (e.g., preliminary or failed experiments that didn't make it into the paper). 
    \end{itemize}
    
\item {\bf Code of ethics}
    \item[] Question: Does the research conducted in the paper conform, in every respect, with the NeurIPS Code of Ethics \url{https://neurips.cc/public/EthicsGuidelines}?
    \item[] Answer: \answerYes{} 
    \item[] Justification: We strictly follow the NeurIPS Code of Ethics.
    \item[] Guidelines:
    \begin{itemize}
        \item The answer \answerNA{} means that the authors have not reviewed the NeurIPS Code of Ethics.
        \item If the authors answer \answerNo, they should explain the special circumstances that require a deviation from the Code of Ethics.
        \item The authors should make sure to preserve anonymity (e.g., if there is a special consideration due to laws or regulations in their jurisdiction).
    \end{itemize}

\item {\bf Broader impacts}
    \item[] Question: Does the paper discuss both potential positive societal impacts and negative societal impacts of the work performed?
    \item[] Answer: \answerYes{} 
    \item[] Justification: Please refer to Appendix~\ref{sec:supp:impacts}.
    \item[] Guidelines:
    \begin{itemize}
        \item The answer \answerNA{} means that there is no societal impact of the work performed.
        \item If the authors answer \answerNA{} or \answerNo, they should explain why their work has no societal impact or why the paper does not address societal impact.
        \item Examples of negative societal impacts include potential malicious or unintended uses (e.g., disinformation, generating fake profiles, surveillance), fairness considerations (e.g., deployment of technologies that could make decisions that unfairly impact specific groups), privacy considerations, and security considerations.
        \item The conference expects that many papers will be foundational research and not tied to particular applications, let alone deployments. However, if there is a direct path to any negative applications, the authors should point it out. For example, it is legitimate to point out that an improvement in the quality of generative models could be used to generate Deepfakes for disinformation. On the other hand, it is not needed to point out that a generic algorithm for optimizing neural networks could enable people to train models that generate Deepfakes faster.
        \item The authors should consider possible harms that could arise when the technology is being used as intended and functioning correctly, harms that could arise when the technology is being used as intended but gives incorrect results, and harms following from (intentional or unintentional) misuse of the technology.
        \item If there are negative societal impacts, the authors could also discuss possible mitigation strategies (e.g., gated release of models, providing defenses in addition to attacks, mechanisms for monitoring misuse, mechanisms to monitor how a system learns from feedback over time, improving the efficiency and accessibility of ML).
    \end{itemize}
    
\item {\bf Safeguards}
    \item[] Question: Does the paper describe safeguards that have been put in place for responsible release of data or models that have a high risk for misuse (e.g., pre-trained language models, image generators, or scraped datasets)?
    \item[] Answer: \answerNA{} 
    \item[] Justification: This paper poses no such risks.
    \item[] Guidelines:
    \begin{itemize}
        \item The answer \answerNA{} means that the paper poses no such risks.
        \item Released models that have a high risk for misuse or dual-use should be released with necessary safeguards to allow for controlled use of the model, for example by requiring that users adhere to usage guidelines or restrictions to access the model or implementing safety filters. 
        \item Datasets that have been scraped from the Internet could pose safety risks. The authors should describe how they avoided releasing unsafe images.
        \item We recognize that providing effective safeguards is challenging, and many papers do not require this, but we encourage authors to take this into account and make a best faith effort.
    \end{itemize}

\item {\bf Licenses for existing assets}
    \item[] Question: Are the creators or original owners of assets (e.g., code, data, models), used in the paper, properly credited and are the license and terms of use explicitly mentioned and properly respected?
    \item[] Answer: \answerYes{} 
    \item[] Justification: We strictly follow the corresponding licenses for existing assets.
    \item[] Guidelines:
    \begin{itemize}
        \item The answer \answerNA{} means that the paper does not use existing assets.
        \item The authors should cite the original paper that produced the code package or dataset.
        \item The authors should state which version of the asset is used and, if possible, include a URL.
        \item The name of the license (e.g., CC-BY 4.0) should be included for each asset.
        \item For scraped data from a particular source (e.g., website), the copyright and terms of service of that source should be provided.
        \item If assets are released, the license, copyright information, and terms of use in the package should be provided. For popular datasets, \url{paperswithcode.com/datasets} has curated licenses for some datasets. Their licensing guide can help determine the license of a dataset.
        \item For existing datasets that are re-packaged, both the original license and the license of the derived asset (if it has changed) should be provided.
        \item If this information is not available online, the authors are encouraged to reach out to the asset's creators.
    \end{itemize}

\item {\bf New assets}
    \item[] Question: Are new assets introduced in the paper well documented and is the documentation provided alongside the assets?
    \item[] Answer: \answerNA{} 
    \item[] Justification: The paper does not release new assets.
    \item[] Guidelines:
    \begin{itemize}
        \item The answer \answerNA{} means that the paper does not release new assets.
        \item Researchers should communicate the details of the dataset\slash code\slash model as part of their submissions via structured templates. This includes details about training, license, limitations, etc. 
        \item The paper should discuss whether and how consent was obtained from people whose asset is used.
        \item At submission time, remember to anonymize your assets (if applicable). You can either create an anonymized URL or include an anonymized zip file.
    \end{itemize}

\item {\bf Crowdsourcing and research with human subjects}
    \item[] Question: For crowdsourcing experiments and research with human subjects, does the paper include the full text of instructions given to participants and screenshots, if applicable, as well as details about compensation (if any)? 
    \item[] Answer: \answerYes{} 
    \item[] Justification: Appendix~\ref{sec:supp:study} describes the task and the three questions asked.
    \item[] Guidelines:
    \begin{itemize}
        \item The answer \answerNA{} means that the paper does not involve crowdsourcing nor research with human subjects.
        \item Including this information in the supplemental material is fine, but if the main contribution of the paper involves human subjects, then as much detail as possible should be included in the main paper. 
        \item According to the NeurIPS Code of Ethics, workers involved in data collection, curation, or other labor should be paid at least the minimum wage in the country of the data collector. 
    \end{itemize}

\item {\bf Institutional review board (IRB) approvals or equivalent for research with human subjects}
    \item[] Question: Does the paper describe potential risks incurred by study participants, whether such risks were disclosed to the subjects, and whether Institutional Review Board (IRB) approvals (or an equivalent approval/review based on the requirements of your country or institution) were obtained?
    \item[] Answer: \answerNo{} 
    \item[] Justification: The study collected only pairwise preference choices on rendered 3D objects, with no personal data and no foreseeable risk to participants.
    \item[] Guidelines:
    \begin{itemize}
        \item The answer \answerNA{} means that the paper does not involve crowdsourcing nor research with human subjects.
        \item Depending on the country in which research is conducted, IRB approval (or equivalent) may be required for any human subjects research. If you obtained IRB approval, you should clearly state this in the paper. 
        \item We recognize that the procedures for this may vary significantly between institutions and locations, and we expect authors to adhere to the NeurIPS Code of Ethics and the guidelines for their institution. 
        \item For initial submissions, do not include any information that would break anonymity (if applicable), such as the institution conducting the review.
    \end{itemize}

\item {\bf Declaration of LLM usage}
    \item[] Question: Does the paper describe the usage of LLMs if it is an important, original, or non-standard component of the core methods in this research? Note that if the LLM is used only for writing, editing, or formatting purposes and does \emph{not} impact the core methodology, scientific rigor, or originality of the research, declaration is not required.
    \item[] Answer: \answerNA{} 
    \item[] Justification: The core method development in this research does not involve LLMs as any
important, original, or non-standard components.
    \item[] Guidelines:
    \begin{itemize}
        \item The answer \answerNA{} means that the core method development in this research does not involve LLMs as any important, original, or non-standard components.
        \item Please refer to our LLM policy in the NeurIPS handbook for what should or should not be described.
    \end{itemize}

\end{enumerate}

\end{document}